# Recursion-Free Online Multiple Incremental/Decremental Analysis Based on Ridge Support Vector Learning

Bo-Wei Chen

*Abstract*—**This study presents a rapid multiple incremental and decremental mechanism based on Weight-Error Curves (WECs) for support-vector analysis. To handle rapidly increasing amounts of data, recursion-free computation is proposed for predicting the Lagrangian multipliers of new samples. This study examines the characteristics of Ridge Support Vector Models, including Ridge Support Vector Machines and Regression, subsequently devising a recursion-free function derived from WECs. With this proposed function, all of the new Lagrangian multipliers can be computed at once without using any gradual step sizes. Moreover, such a function can relax a constraint, where the increment of new multiple Lagrangian multipliers should be the same in the previous work, thereby easily satisfying the requirement of Karush–Kuhn–Tucker (KKT) conditions. The proposed mechanism no longer requires typical time-consuming bookkeeping strategies, which compute the step size by checking all the training samples in each incremental round. Experiments were carried out on open datasets for evaluating our work. The results showed that the computational speed was successfully enhanced, better than the baselines. Besides, the accuracy still remained. These findings revealed that the proposed method was appropriate for incremental/decremental learning, thereby demonstrating the effectiveness of the proposed idea.**

## I. Introduction

THIS study examines efficient classification for big streams — A type of large-scale and real-time streaming data that require high speed and large bandwidth. With the advancement of Internet of Things (IoTs), the source nodes of a wireless sensor network can rapidly collect and transmit information to a fusion center [1], or a sink node which is designed for data aggregation. However, large-scale sensornets with massive amounts of streaming deplete the computation resource of sink nodes, either for the storage or the processing unit.

For dealing with large amounts of data, typical solutions involve distributed processing and incremental analysis. The former relies on divide-and-conquer strategies, where a larger dataset is separated into smaller subsets and subsequently processed by independent machines in parallel. The solution to the original problem is obtained by merging suboptimal solutions generated from smaller subsets. MapReduce is a famous example of frameworks implemented in "Apache Hadoop" for distributed processing. Although divide-and-conquer strategies are applicable to big data, however, the training dataset should be fixed. This means when new samples arrive, the entire system must perform the training procedure again. Otherwise, a hybrid mode integrates both distributed processing and incremental analysis is used instead.

Unlike distributed processing, incremental analysis allows systems to add new training samples without retraining. Furthermore, incremental analysis also supports single-instance training, multiple-instance training, or both of them. They are all conducive to relief of computational loads as earlier calculation results can be reserved for updating the new system in the future. For multiple-instance training, or batch incremental analysis, it is useful for big data. As the size of data is far beyond the capability of one machine, especially when the memory space cannot accommodate the entire data at once, the entire set can be divided into several batches.

Recently, many efforts [2-6] have been devoted to incremental classification, for instance, Kernel Ridge Regression (KRR) [7] and Support Vector Machines (SVMs). This study particularly concentrates on the SVM due to its efficiency in training. Literature reviews [7] showed that the complexity of KRR was as high as $O(N^3)$, whereas that of SVMs was merely $O(N^2)$, where $N$ denotes the number of samples.

In 2000, Cauwenberghs and Poggio [2] established a milestone for incremental SVMs as they discovered the equilibrium between old Lagrangian multipliers and new ones. A differential form was derived from the cost function of SVMs and the Karush–Kuhn–Tucker (KKT) [3] conditions. Such a differential form supported single incremental and decremental learning. The derivation was shown in a subsequent study [4]. A recursive procedure was introduced to update the matrix formed by the original support vectors and the kernel matrix when a single instance was changed. The authors also devised a strategy called "bookkeeping," or the accounting strategy mentioned in [5], to determine the largest increment/decremental amount of existing Lagrangian

B.-W. Chen is with the School of Information Technology, Monash University, Australia (dennisbwc@gmail.com; bo-wei.chen@monash.edu)

Table I
Characteristics of Incremental/Decremental Mechanisms

| Method | Single Incremental/Decremental | Multiple Incremental/Decremental | Bookkeeping | Step Size for Multiple Incremental/Decremental |
|---|---|---|---|---|
| Cauwenberghs & Poggio [2] | Yes | No | Necessary | N/A |
| Martin [8] & Ma *et al*. [9] | Yes | No | Necessary | N/A |
| Laskov *et al*. [5] | Yes | No | Necessary | N/A |
| Karasuyama & Takeuchi [6] | Yes | Yes | Necessary | Recursion |
| Proposed | Yes | Yes | Unnecessary | All at Once |

multipliers. The model by Cauwenberghs and Poggio has inspired subsequent studies, for example [4], [5], and [6]. Laskov *et al*. [5] summarized the methodology developed in [2] by presenting a systematic analytical solution. Such a solution explicitly and clearly elaborated the changes in Lagrangian multipliers with respect to three cases: Unbounded support vectors, bounded support vectors, and non-support vectors. Each vector corresponded to one Lagrangian multiplier. Furthermore, they also presented recursive matrix updates and matrix decomposition that were conducive to incremental/decremental matrix computation. Karasuyama and Takeuchi [6] advanced the approach proposed by [2] and developed a strategy for multiple incremental/decremental learning. Multiple incremental and decremental processing were combined together during the update of the system, without being executed separately. Karasuyama and Takeuchi simplified the bookkeeping strategy mentioned in [2] by searching the shortest and easiest path when the Lagrangian multipliers of new samples were computed. The definition of the path in their work represented a series of incremental/decremental changes in new Lagrangian multipliers. The system required recursive computation for a proper incremental step-size. As the Lagrangian multipliers of all the new samples have the same incremental amount, it is time-consuming to satisfy KKT conditions at once. The common part in [2], [5], and [6] is that the incremental amount was absorbed by the Lagrangian multipliers of existing unbounded support vectors and the bias term to maintain an equilibrium. Even when the existing samples in their original regions (i.e., within the margin, on the margin, and outside the margin) are moved into the other regions, their Lagrangian multipliers never exceed the constraints during the update.

The concept of incremental SVMs was also applicable to Support Vector Regression (SVR) as SVR is a variation of SVMs. Based on the same differential form, Martin [8] and Ma *et al*. [9] devised an incremental/decremental model for SVR with several modifications, of which the mechanism supported single-instance online learning. Like [4-6], both [8] and [9] required a bookkeeping strategy to estimate appropriate step sizes for the Lagrange multiplier of a new sample. The membership of existing samples could be changed after the update.

In addition to the above-mentioned works [2, 5, 6, 8, 9], some studies focused on proximal solutions, for example, [10-12]. The system in [10] examined locality of samples based on Radial Basis Functions (RBFs). Following locality checking, the system reestimated the weight of the existing training data near the new instances by using an adjusting function. Fung and Mangasarian devised a Proximal SVM (PSVM) based on linear kernels for rapid incremental learning [11, 13]. The PSVM could adjust the hyperplane by pushing apart two parallel marginal planes in an efficient way. Thus, the PSVM was capable of processing large-scale data compared with the state-of-the-art techniques according to the experiments. Tveit and Engum [12] furthered the work by Fung and Mangasarian by proposing a new data structure to speed up the computation. In summary, the merit of proximal solutions is fast training and low computational burdens. Additionally, proximal methods are suitable for applications that are less sensitive to prediction accuracy. However, when more new instances arrive, biased estimation increases. Therefore, optimal solutions are favored in most applications.

The contributions of this study are summarized as follows.

- We discovered the functions of WECs for Ridge SVMs and Ridge SVR, which are used for prediction of Lagrangian multipliers based on errors.
- The Lagrangian multipliers of new samples are directly predicted all at once based on WECs. No recursive computation for determining these Lagrangian multipliers is required.
- Typical bookkeeping strategies [2, 5, 6, 8, 9] are no longer necessary. Bookkeeping strategies compute the minimum step size by checking all the training samples in each incremental round. This consumes too much computational time.
- Different values are assigned to the Lagrangian multipliers of new samples, subsequently easily satisfying the requirement of KKT conditions.
- No prior analysis on data distribution is required to analyze the weighting vector of SVMs and SVR.
- We advance single incremental/decremental SVR to multiple incremental/decremental SVR.
- Ridge parameters generate a more flexible region during sequential minimal optimization, which typical SVMs and SVR do not have. Besides, ridge parameters prevent kernel matrices from violating Mercer conditions.

Table I compares the prior works [2, 5, 6, 8, 9] and ours. These five works are all based on the differential equation by Cauwenberghs and Poggio [2]. Only our work and the system by Karasuyama and Takeuchi [6] support multiple incremental and decremental learning.

The rest of this paper is organized as follows. Section II first introduces the concept of Ridge SVMs. Subsequently, typical single incremental/decremental learning is introduced in Section III, and Section IV describes the details of the proposed multiple incremental/decremental learning. Section V further extends the proposed idea to Ridge SVR. Section VI shows experimental results. Conclusions are finally drawn in Section VII.

## II. Ridge Support Vector Machines

This study begins with Ridge SVMs, where a ridge parameter $\rho$ is imposed on the kernel matrix. Let $\mathbf{x}_i$ represent an $M$-dimensional training sample and $y_i$ denote its label, where $i = 1,\ldots,N$. The objective of SVMs is maximizing the separation margin. Thus, the cost function of SVMs is defined as

$$\min_{\mathbf{u},b,\xi}\left\{\frac{1}{2}\|\mathbf{u}\|^2 + C\sum_{i}^{N}\xi_i\right\} \tag{1}$$

$$\text{s.t.}\begin{cases} y_i\epsilon_i + \xi_i \geq 0 \\ \epsilon_i = \mathbf{u}^{\mathrm{T}}\phi(x_i) + b - y_i \\ \xi_i \geq 0 \end{cases}$$

where $\phi(\mathbf{x}_i)$ denotes the intrinsic-space feature vector of $\mathbf{x}_i$, $\mathbf{u}$ is the weight vector, and $b$ is a bias term. Furthermore, $C$ is the penalty, $\xi$ represents a slack variable, and $\epsilon$ denotes the classification error. By applying Lagrange multipliers α, the original problem is converted to the dual-optimization problem, which is equivalent to the following Wolfe dual formulation [1].

$$\max_{\alpha}\mathfrak{L}_{\text{SVM}}(\alpha) = \max_{\alpha}\left\{\sum_{i=1}^{N}\alpha_i - \frac{1}{2}\sum_{i=1}^{N}\sum_{j=1}^{N}\alpha_i\alpha_j y_i y_j K\left(\mathbf{x}_i,\mathbf{x}_j\right)\right\} \tag{2}$$

$$\text{s.t.}\begin{cases} \sum_{i=1}^{N}\alpha_i y_i = 0 \\ 0 \leq \alpha_i \leq C \end{cases}$$

where $i$ and $j$ respectively specify the index of a sample, and $K(\mathbf{x}_i,\mathbf{x}_j)$ is a kernel function that calculates the distance between $\mathbf{x}_i$ and $\mathbf{x}_j$ in the empirical space.

When rewritten in a matrix form, (2) becomes

$$\max_{\mathbf{a}}\left\{\mathbf{a}^{\mathrm{T}}\mathbf{y} - \frac{1}{2}\mathbf{a}^{\mathrm{T}}\mathbf{K}\mathbf{a}\right\} \tag{3}$$

$$\text{s.t.}\begin{cases} \sum_{i=1}^{N}a_i = 0 \\ 0 \leq a_i y_i \leq C \end{cases}$$

where $\alpha_i = a_i \times y_i$. Notably, $\mathbf{K}$ is a kernel matrix. To regularize the above-mentioned model, this study uses a ridge parameter $\rho$ to control the formation of $\mathbf{K}$. Accordingly, the original problem (3) subsequently becomes

$$\max_{\mathbf{a}}\left\{\mathbf{a}^{\mathrm{T}}\mathbf{y} - \frac{1}{2}\mathbf{a}^{\mathrm{T}}\left(\mathbf{K} + \rho\mathbf{I}\right)\mathbf{a}\right\} \tag{4}$$

$$\text{s.t.}\begin{cases} \sum_{i=1}^{N}a_i = 0 \\ 0 \leq a_i y_i \leq C \end{cases}.$$

The ridge parameter has two advantages. One is to provide a mechanism that rotates the hyperplane. When combined with $C$ and $\xi$, Ridge Support Vector Machines generate a more flexible hyperplane than classic Support Vector Machines. Second, the ridge parameter prevents the kernel matrix from violating Mercer conditions.

The effect of the ridge parameter is reflected in the Weight-Error Curve (WEC), which is generated by plotting Lagrange multipliers α against the corresponding classification errors. In typical Support Vector Machines, WECs usually exhibit the shapes like Fig. 1, where the horizontal axis specifies errors, and the corresponding Lagrange multipliers are displayed in the vertical axis. Blue and red marks respectively represent positive and negative classes. A vertical transitional region usually exists between two horizontal lines. Such a transition region contains the Lagrange multipliers of all the unbounded support vectors. Recall that the Lagrange multipliers of unbounded support vectors are $0 < \alpha < C$ with the classification error equal to zero.

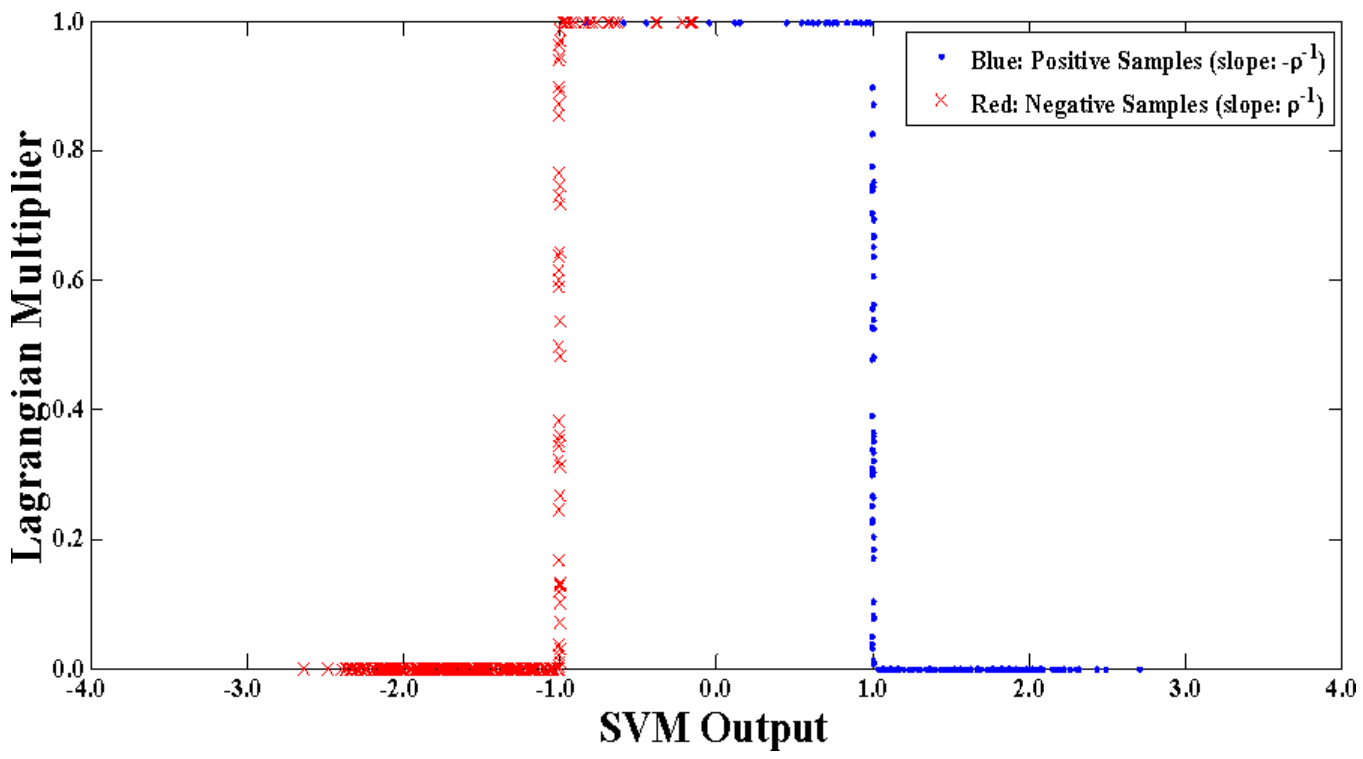

Fig. 1. WEC of a typical SVM, where $\rho$ = 0.0 and $C$ = 1.0. Notably, when the ridge parameter equals zero, Ridge SVMs are equivalent to typical SVMs. The left curve is the result of positive training samples, whereas the right one is that of negative training samples. For clarity, we use SVM outputs instead of errors in this figure.

Compared with the WECs of Support Vector Machines, those of Ridge Support Vector Machines exhibit a ramp in the transitional region, as shown in Fig. 2. This ramp accommodates more unbounded support vectors as there is a slope in that region. In other words, when the algorithm searches for the solutions to Lagrange multipliers, the ramp relaxes constraints limited by KKT conditions.

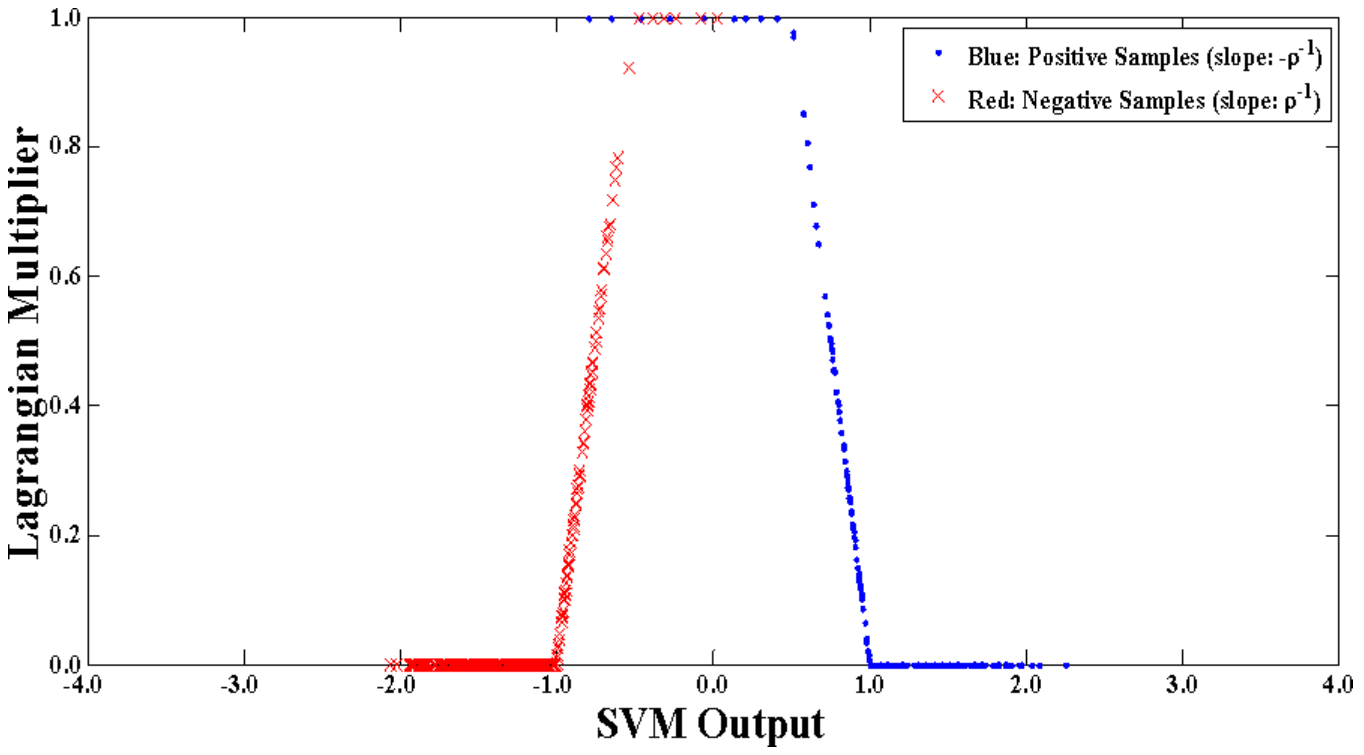

Fig. 2. WEC of a Ridge SVM, where $\rho$ = 0.5 and $C$ = 1.0. For clarity, we use SVM outputs instead of errors in this figure.

We respectively model the equations corresponding to these two curves as follows.

For positive samples in the slope on the right side,

$$\alpha_i = \rho - \rho^{-1} f(\mathbf{x}_i), \quad \text{s.t. } y_i f(\mathbf{x}_i) \geq 1. \tag{5}$$

For negative samples in the left slope,

$$\alpha_i = \rho + \rho^{-1} f(\mathbf{x}_i), \quad \text{s.t. } y_i f(\mathbf{x}_i) \leq -1. \tag{6}$$

Furthermore, the shift of the two functions is also $\rho$.

## III. Single Incremental/Decremental Leaning

This section begins with the work by Cauwenberghs and Poggio [2], where an important single incremental equation was introduced. Such an equation is the central idea of [2, 5, 6] and our work.

To illustrate the idea of [2], the Wolfe dual formulation with a ridge parameter in (2) can be rewritten as

$$\begin{aligned}
&\max_{\alpha} \mathfrak{L}_{\text{SVM}}(\alpha) \\
&= \max_{\alpha} \left\{ \sum_{i=1}^{N} \alpha_i - \frac{1}{2} \sum_{i=1}^{N} \sum_{j=1}^{N} \alpha_i \alpha_j y_i y_j \left( K(\mathbf{x}_i, \mathbf{x}_j) + \rho \mathbf{I}_{i,j} \right) \right\} \\
&= \max_{\alpha} \left\{ \sum_{i=1}^{N} \alpha_i - \frac{1}{2} \sum_{i=1}^{N} \sum_{j=1}^{N} \alpha_i \alpha_j y_i y_j \left( K(\mathbf{x}_i, \mathbf{x}_j) + \rho \mathbf{I}_{i,j} \right) - b \sum_{i=1}^{N} y_i \alpha_i \right\}
\end{aligned} \tag{7}$$

where the last term is put back into the equation. The upper part and the lower part are equivalent because $\sum_{i=1}^{N} \alpha_i y_i = 0$ when we take the partial derivative with respect $b$. For consistency with [2, 5, 6], (7) is subsequently rewritten as

$$\begin{aligned}
&\max_{\alpha} \mathfrak{L}_{\text{SVM}}(\alpha) \\
&\equiv \min_{\alpha} \mathfrak{H}_{\text{SVM}}(\alpha) \\
&= \min_{\alpha} \left\{ \frac{1}{2} \sum_{i=1}^{N} \sum_{j=1}^{N} \alpha_i \alpha_j y_i y_j \left( K(\mathbf{x}_i, \mathbf{x}_j) + \rho \mathbf{I}_{i,j} \right) - \sum_{i=1}^{N} \alpha_i + b \sum_{i=1}^{N} y_i \alpha_i \right\} \\
&= \min_{\alpha} \left\{ \frac{1}{2} \sum_{i=1}^{N} \sum_{j=1}^{N} \alpha_i \alpha_j Q_{ij} - \sum_{i=1}^{N} \alpha_i + b \sum_{i=1}^{N} y_i \alpha_i \right\}
\end{aligned} \tag{8}$$

where $\mathbf{Q}$ represents $\mathbf{y}^{\mathrm{T}}\mathbf{y} \otimes (\mathbf{K} + \rho\mathbf{I})$, and $\otimes$ denotes Hadamard product operator.

As $\alpha_i$ and $b$ are unknown, taking the partial derivative yields

$$\frac{\partial \mathfrak{H}_{\text{SVM}}}{\partial \alpha_i} = \sum_{j}^{N} Q_{ij} \alpha_j + y_i b - 1 \tag{9}$$

and

$$\frac{\partial \mathfrak{H}_{\text{SVM}}}{\partial b} = \sum_{j}^{N} y_j \alpha_j = 0. \tag{10}$$

Equation (10) is the Orthogonal-Hyperplane Property mentioned in [7] when the system satisfies the KKT conditions.

Let $\mathfrak{G}_i$ represent (9). Subsequently,

$$\begin{aligned}
\mathfrak{G}_i &= \sum_{j}^{N} Q_{ij} \alpha_j + y_i b - 1 \\
&= \sum_{s \in \mathcal{S}} Q_{is} \alpha_s + \sum_{\iota \in \mathcal{I}} Q_{i\iota} \alpha_\iota + \sum_{o \in \mathcal{O}} Q_{io} \alpha_o + y_i b - 1 \\
&= \sum_{s \in \mathcal{S}} Q_{is} \alpha_s + \sum_{\iota \in \mathcal{I}} Q_{i\iota} \alpha_\iota + y_i b - 1
\end{aligned} \tag{11}$$

where $\mathcal{S}$ stands for unbounded support vectors, $\mathcal{I}$ represents bounded support vectors, and $\mathcal{O}$ denotes nonsupport vectors. The relation between $\mathfrak{G}$ and the Lagrangian multipliers of $\mathcal{S}$, $\mathcal{I}$, and $\mathcal{O}$ are respectively

$$\begin{cases}
\mathfrak{G}_i = 0, & 0 < \alpha_i < C, \quad i \in \mathcal{S} \\
\mathfrak{G}_i < 0, & \alpha_i = C, \quad i \in \mathcal{I} \\
\mathfrak{G}_i > 0, & \alpha_i = 0, \quad i \in \mathcal{O}
\end{cases}. \tag{12}$$

Denote $\mathbf{x}_d$ as a new training instance. Subsequently, a single incremental equation is derived based on [2] and [4]:

$$\Delta \mathfrak{G}_i = Q_{id} \Delta \alpha_d + \sum_{s \in \mathcal{S}} Q_{is} \Delta \alpha_s + y_i \Delta b \tag{13}$$

and

$$\mathfrak{D} = y_d \Delta\alpha_d + \sum_{s\in\mathcal{S}} y_s \Delta\alpha_s \ .$$

Additionally, to maintain the equilibrium of the system, the following properties should be satisfied. That is,

$$\Delta\mathfrak{G}_i = Q_{id}\Delta\alpha_d + \sum_{s\in\mathcal{S}} Q_{is}\Delta\alpha_s + y_i \Delta b = 0, \quad i\in\mathcal{S}, \tag{14}$$

and

$$\mathfrak{D} = y_d \Delta\alpha_d + \sum_{s\in\mathcal{S}} y_s \Delta\alpha_s = 0 \tag{15}$$

where the second equation is derived from the Orthogonal-Hyperplane Property [7], i.e.,

$$\sum_{i}^{N} y_i \alpha_i + y_d \alpha_d = 0 \ . \tag{16}$$

Based on (14) and (15), any increment on the Lagrangian multiplier of a new training instance is fed backward to the existing Lagrangian multipliers in $\mathcal{S}$ and $b$. The change of the existing Lagrangian multipliers in $\mathcal{S}$ and $b$ absorbs the increment of the new Lagrangian multiplier.

## IV. Proposed Multiple Incremental/Decremental Leaning

This section firstly introduces the method by Karasuyama and Takeuchi [6], where recursive computation was developed for multiple incremental and decremental analyses. Subsequently, the proposed rapid support vector analysis based on Ridge SVMs is integrated to the method by Karasuyama and Takeuchi, so that the system no longer requires recursive computation.

Assume $\mathcal{D}$ represents the set of new training samples. Also assume $\mathcal{R}$ denotes the set of existing training samples that are about to be removed. The idea in [6] is to seek the shortest path during the analysis. That is, for the incremental phase, all the Lagrangian multipliers of new training samples have the same increment. Additionally, the direction of the increment is from zero to $C$. This implies that the direction is diagonal in the space. For the decremental phase, all the Lagrangian multipliers in $\mathcal{R}$ are directly reduced to zero.

The original approach required a step size $\eta$ for incremental and decremental analyses, as shown in (17) and (18). Therefore, a bookkeeping strategy in (26) is necessary.

$$\Delta\boldsymbol{\alpha}_{\mathcal{D}} = \eta\left(C\mathbf{1} - \boldsymbol{\alpha}_{\mathcal{D}}\right) \tag{17}$$

and

$$\Delta\boldsymbol{\alpha}_{\mathcal{R}} = -\eta\boldsymbol{\alpha}_{\mathcal{R}} \ . \tag{18}$$

The decremental process needs a step size simply because we perform the incremental process at the same time.

Based on multiple incremental and decremental equations in [6], (14) and (15) can be rewritten as

$$\begin{aligned}\Delta\mathfrak{G}_i &= \sum_{d\in\mathcal{D}} Q_{id}\Delta\alpha_d + \sum_{r\in\mathcal{R}} Q_{ir}\Delta\alpha_r + \sum_{s\in\mathcal{S}} Q_{is}\Delta\alpha_s + y_i\Delta b \\ &= 0,\ i\in\mathcal{S}\end{aligned} \tag{19}$$

and

$$\sum_{d\in\mathcal{D}} y_d\Delta\alpha_d + \sum_{r\in\mathcal{R}} y_r\Delta\alpha_r + \sum_{s\in\mathcal{S}} y_s\Delta\alpha_s = 0 \ . \tag{20}$$

Equations (19) and (20) can be converted into a matrix form as follows.

$$\begin{bmatrix} 0 & \mathbf{y}_{\mathcal{S}}^{\mathrm{T}} \\ \mathbf{y}_{\mathcal{S}} & \mathbf{Q}_{\mathcal{S}} \end{bmatrix}\begin{bmatrix} \Delta b \\ \Delta\boldsymbol{\alpha}_{\mathcal{S}} \end{bmatrix} + \begin{bmatrix} \mathbf{y}_{\mathcal{D}}^{\mathrm{T}} & \mathbf{y}_{\mathcal{R}}^{\mathrm{T}} \\ \mathbf{Q}_{\mathcal{S},\mathcal{D}} & \mathbf{Q}_{\mathcal{S},\mathcal{R}} \end{bmatrix}\begin{bmatrix} \Delta\boldsymbol{\alpha}_{\mathcal{D}} \\ \Delta\boldsymbol{\alpha}_{\mathcal{R}} \end{bmatrix} = 0 \ . \tag{21}$$

Rearranging the equation and substitution (29) and (30) into (31) yield

$$\begin{aligned}\begin{bmatrix} \Delta b \\ \Delta\boldsymbol{\alpha}_{\mathcal{S}} \end{bmatrix} &= -\begin{bmatrix} 0 & \mathbf{y}_{\mathcal{S}}^{\mathrm{T}} \\ \mathbf{y}_{\mathcal{S}} & \mathbf{Q}_{\mathcal{S}} \end{bmatrix}^{-1}\begin{bmatrix} \mathbf{y}_{\mathcal{D}}^{\mathrm{T}} & \mathbf{y}_{\mathcal{R}}^{\mathrm{T}} \\ \mathbf{Q}_{\mathcal{S},\mathcal{D}} & \mathbf{Q}_{\mathcal{S},\mathcal{R}} \end{bmatrix}\begin{bmatrix} \Delta\boldsymbol{\alpha}_{\mathcal{D}} \\ \Delta\boldsymbol{\alpha}_{\mathcal{R}} \end{bmatrix} \\ &= -\eta\begin{bmatrix} 0 & \mathbf{y}_{\mathcal{S}}^{\mathrm{T}} \\ \mathbf{y}_{\mathcal{S}} & \mathbf{Q}_{\mathcal{S}} \end{bmatrix}^{-1}\begin{bmatrix} \mathbf{y}_{\mathcal{D}}^{\mathrm{T}} & \mathbf{y}_{\mathcal{R}}^{\mathrm{T}} \\ \mathbf{Q}_{\mathcal{S},\mathcal{D}} & \mathbf{Q}_{\mathcal{S},\mathcal{R}} \end{bmatrix}\begin{bmatrix} C\mathbf{1}-\boldsymbol{\alpha}_{\mathcal{D}} \\ -\boldsymbol{\alpha}_{\mathcal{R}} \end{bmatrix}\end{aligned} \tag{22}$$

and

$$\begin{aligned}\mathbf{y}\otimes\Delta\mathbf{f} &= \eta\boldsymbol{\varphi} \\ &= \eta\left(\begin{bmatrix} \mathbf{y} \\ \mathbf{Q}_{:,\mathcal{S}} \end{bmatrix}^{\mathrm{T}}\begin{bmatrix} 0 & \mathbf{y}_{\mathcal{S}}^{\mathrm{T}} \\ \mathbf{y}_{\mathcal{S}} & \mathbf{Q}_{\mathcal{S}} \end{bmatrix}^{-1}\begin{bmatrix} \mathbf{y}_{\mathcal{D}}^{\mathrm{T}} & \mathbf{y}_{\mathcal{R}}^{\mathrm{T}} \\ \mathbf{Q}_{\mathcal{S},\mathcal{D}} & \mathbf{Q}_{\mathcal{S},\mathcal{R}} \end{bmatrix}\right. \\ &\quad \left. + \begin{bmatrix} \mathbf{Q}_{:,\mathcal{D}} \\ \mathbf{Q}_{:,\mathcal{R}} \end{bmatrix}^{\mathrm{T}}\right)\begin{bmatrix} C\mathbf{1}-\boldsymbol{\alpha}_{\mathcal{D}} \\ -\boldsymbol{\alpha}_{\mathcal{R}} \end{bmatrix}.\end{aligned} \tag{23}$$

To maintain the equilibrium, i.e., to absorb the change in $\mathcal{D}$ and $\mathcal{R}$ by using $\Delta\boldsymbol{\alpha}_{\mathcal{S}}$ and $\Delta b$, the condition in $\mathcal{S}$ should always hold. Thus,

$$\begin{cases} y_i \cdot f_i = 1 \\ y_i \cdot \Delta f_i = 0 \end{cases} \Leftrightarrow y_i\left(f_i + \Delta f_i\right) = 1, \quad i\in\mathcal{S} \ . \tag{24}$$

The movement of $\mathcal{D}$ is from $\mathcal{O}$ to $\mathcal{S}$ and, if necessary, finally reaches $\mathcal{I}$. The movement of $\mathcal{R}$ directly arrive in $\mathcal{O}$, which is not affected by the step size $\eta$. The maximum change in Lagrangian multipliers in $\mathcal{D}$ and $\mathcal{R}$ is $\boldsymbol{\varphi}$, regulated by $\eta$.

$$y_i \cdot \Delta f_i = 1 - y_i \cdot f_i = -\mathfrak{G}_i = \eta\varphi_i \ . \tag{25}$$

To ensure the Lagrangian multipliers of existing samples in $\mathcal{S}$ fully absorb the change in $\mathcal{D}$ and $\mathcal{R}$ without violating any constraints, i.e., $0 \le \alpha \le C$, the minimum step size is selected based on

$$\hat{\eta} = \arg\min_i \left\| -\mathfrak{G}_i / \varphi_i \right\|, \quad i \in \mathcal{S} \cup \mathcal{I} \cup \mathcal{O}. \tag{26}$$

Such a procedure is called bookkeeping in [2, 5, 6, 8, 9]. Notably, Lagrangian multipliers in $\mathcal{S}$ are allowed to become zero or $C$. In such cases, membership is accordingly changed after the minimum step size is decided.

The update of the step size $\eta$ involves (22), (23), and (26). It recursively updates $\eta$ until $\eta$ reaches one or converges. For each iteration, the system needs to checks (26) for all the training and new samples, so that a minimum step size is selected. Moreover, the system needs to compute (22) to maintain the equilibrium. However, such an update proposed by Karasuyama and Takeuchi created too much complexity when data were frequently changed. Moreover, recursion also consumed computational time.

To rapidly determine the Lagrangian multipliers of new training samples, (5) and (6) are introduced herein to replace (17). The proposed method no longer relies on the selection of step sizes $\eta$.

For new positive training samples,

$$\alpha_d = \rho - \rho^{-1} f(\mathbf{x}_d), \tag{27}$$

$$\text{s.t.} \begin{cases} y_d f(\mathbf{x}_d) \ge 1 \\ \alpha_d = C, & \text{if } \alpha_d > C \\ \alpha_d = 0, & \text{if } \alpha_d < 0 \end{cases}.$$

For new negative training samples,

$$\alpha_d = \rho + \rho^{-1} f(\mathbf{x}_d), \tag{28}$$

$$\text{s.t.} \begin{cases} y_d f(\mathbf{x}_d) \le -1 \\ \alpha_d = C, & \text{if } \alpha_d > C \\ \alpha_d = 0, & \text{if } \alpha_d < 0 \end{cases}.$$

To avoid generating inappropriate Lagrangian multipliers, the system rechecks (12) after Lagrangian multipliers are predicted by using (27) and (28). Any values that violate (12) are reset to zero.

To simplify the notation in (27) and (28), $\Omega(\cdot)$ is used to represent the equations.

$$\Delta\boldsymbol{\alpha}_{\mathcal{D}} = \Omega(\mathbf{x}_{\mathcal{D}}). \tag{29}$$

As no step size is required, therefore,

$$\Delta\boldsymbol{\alpha}_{\mathcal{R}} = -\boldsymbol{\alpha}_{\mathcal{R}}. \tag{30}$$

Rearranging the equation and substitution (29) and (30) into (31) yield

$$\begin{aligned} \begin{bmatrix} \Delta b \\ \Delta\boldsymbol{\alpha}_{\mathcal{S}} \end{bmatrix} &= -\begin{bmatrix} 0 & \mathbf{y}_{\mathcal{S}}^{\mathrm{T}} \\ \mathbf{y}_{\mathcal{S}} & \mathbf{Q}_{\mathcal{S}} \end{bmatrix}^{-1} \begin{bmatrix} \mathbf{y}_{\mathcal{D}}^{\mathrm{T}} & \mathbf{y}_{\mathcal{R}}^{\mathrm{T}} \\ \mathbf{Q}_{\mathcal{S},\mathcal{D}} & \mathbf{Q}_{\mathcal{S},\mathcal{R}} \end{bmatrix} \begin{bmatrix} \Delta\boldsymbol{\alpha}_{\mathcal{D}} \\ \Delta\boldsymbol{\alpha}_{\mathcal{R}} \end{bmatrix} \\ &= -\begin{bmatrix} 0 & \mathbf{y}_{\mathcal{S}}^{\mathrm{T}} \\ \mathbf{y}_{\mathcal{S}} & \mathbf{Q}_{\mathcal{S}} \end{bmatrix}^{-1} \begin{bmatrix} \mathbf{y}_{\mathcal{D}}^{\mathrm{T}} & \mathbf{y}_{\mathcal{R}}^{\mathrm{T}} \\ \mathbf{Q}_{\mathcal{S},\mathcal{D}} & \mathbf{Q}_{\mathcal{S},\mathcal{R}} \end{bmatrix} \begin{bmatrix} \Omega(\mathbf{x}_{\mathcal{D}}) \\ -\boldsymbol{\alpha}_{\mathcal{R}} \end{bmatrix} \end{aligned}. \tag{31}$$

Based on the Schur complement theory,

$$\begin{bmatrix} 0 & \mathbf{y}_{\mathcal{S}}^{\mathrm{T}} \\ \mathbf{y}_{\mathcal{S}} & \mathbf{Q}_{\mathcal{S}} \end{bmatrix}^{-1} = \begin{bmatrix} z & -z\mathbf{y}_{\mathcal{S}}^{\mathrm{T}}\mathbf{Q}_{\mathcal{S}}^{-1} \\ -z\mathbf{Q}_{\mathcal{S}}^{-1}\mathbf{y}_{\mathcal{S}} & z\mathbf{Q}_{\mathcal{S}}^{-1}\mathbf{y}_{\mathcal{S}}\mathbf{y}_{\mathcal{S}}^{\mathrm{T}}\mathbf{Q}_{\mathcal{S}}^{-1} + \mathbf{Q}_{\mathcal{S}}^{-1} \end{bmatrix} \tag{32}$$

where $z$ is a scalar derived from the inverse of $\mathbf{Q}_{\mathcal{S}}$ shown as follows:

$$z = -\left(\mathbf{y}_{\mathcal{S}}^{\mathrm{T}}\mathbf{Q}_{\mathcal{S}}^{-1}\mathbf{y}_{\mathcal{S}}\right)^{-1} \tag{33}$$

and

$$\mathbf{Q}_{\mathcal{S}} = \mathbf{y}_{\mathcal{S}}^{\mathrm{T}}\mathbf{y}_{\mathcal{S}} \otimes \left(\mathbf{K}_{\mathcal{S}} + \rho\mathbf{I}\right).$$

To support computing $\mathbf{Q}^{-1}[\ell+1]$ based on the previous result $\mathbf{Q}^{-1}[\ell]$, where $\ell$ denotes iterations, the system can use the method in the appendix for acceleration. For empty $\mathcal{S}$, the system is required to select two new training samples to rebuild $\mathcal{S}$ based on a closed-form equation in [5].

## V. Multiple Incremental/Decremental Ridge Support Vector Regression

This section extends the idea of Ridge SVMs to Ridge SVR. In original SVR [14], each sample must satisfy two optimization constraints with respect to slack variables, $\xi$ and $\xi^*$. Therefore,

$$\min_{\mathbf{u},b,\xi,\xi_i^*} \left\{ \frac{1}{2}\|\mathbf{u}\|^2 + C\sum_i^N \left(\xi_i + \xi_i^*\right) \right\} \tag{34}$$

$$\text{s.t.} \begin{cases} \epsilon_i \le \varepsilon + \xi_i \\ -\epsilon_i \le \varepsilon + \xi_i^* \\ \epsilon_i = \mathbf{u}^{\mathrm{T}}\phi(x_i) + b - y_i \\ \xi_i, \xi_i^* \ge 0 \end{cases}.$$

This leads to the Wolfe dual formulation [15]. That is,

$$\min_{\alpha,\alpha^*} \mathfrak{H}_{\mathrm{SVR}}\left(\alpha,\alpha^*\right) = \min_{\alpha,\alpha^*}\left\{\frac{1}{2}\sum_{i=1}^{N}\sum_{j=1}^{N}\left(\alpha_i-\alpha_i^*\right)\left(\alpha_j-\alpha_j^*\right)\mathcal{Q}_{ij} - \varepsilon\sum_{i=1}^{N}\left(\alpha_i+\alpha_i^*\right)+\sum_{i=1}^{N}y_i\left(\alpha_i-\alpha_i^*\right)\right\} \quad (35)$$

$$\text{s.t.}\begin{cases}0\le\alpha_i,\alpha_i^*\le C\\ \sum_{i}^{N}\left(\alpha_i-\alpha_i^*\right)=0\end{cases}.$$

where $\mathcal{Q}$ represents $\mathbf{K}+\rho\mathbf{I}$. Besides, α and α* are Lagrangian multipliers introduced to resolve the constraints. When $\rho \neq 0$, SVR becomes Ridge SVR. Like Ridge SVMs, the difference in WECs between SVR and Ridge SVR lies in the ramp, as shown in Fig. 3 and Fig. 4.

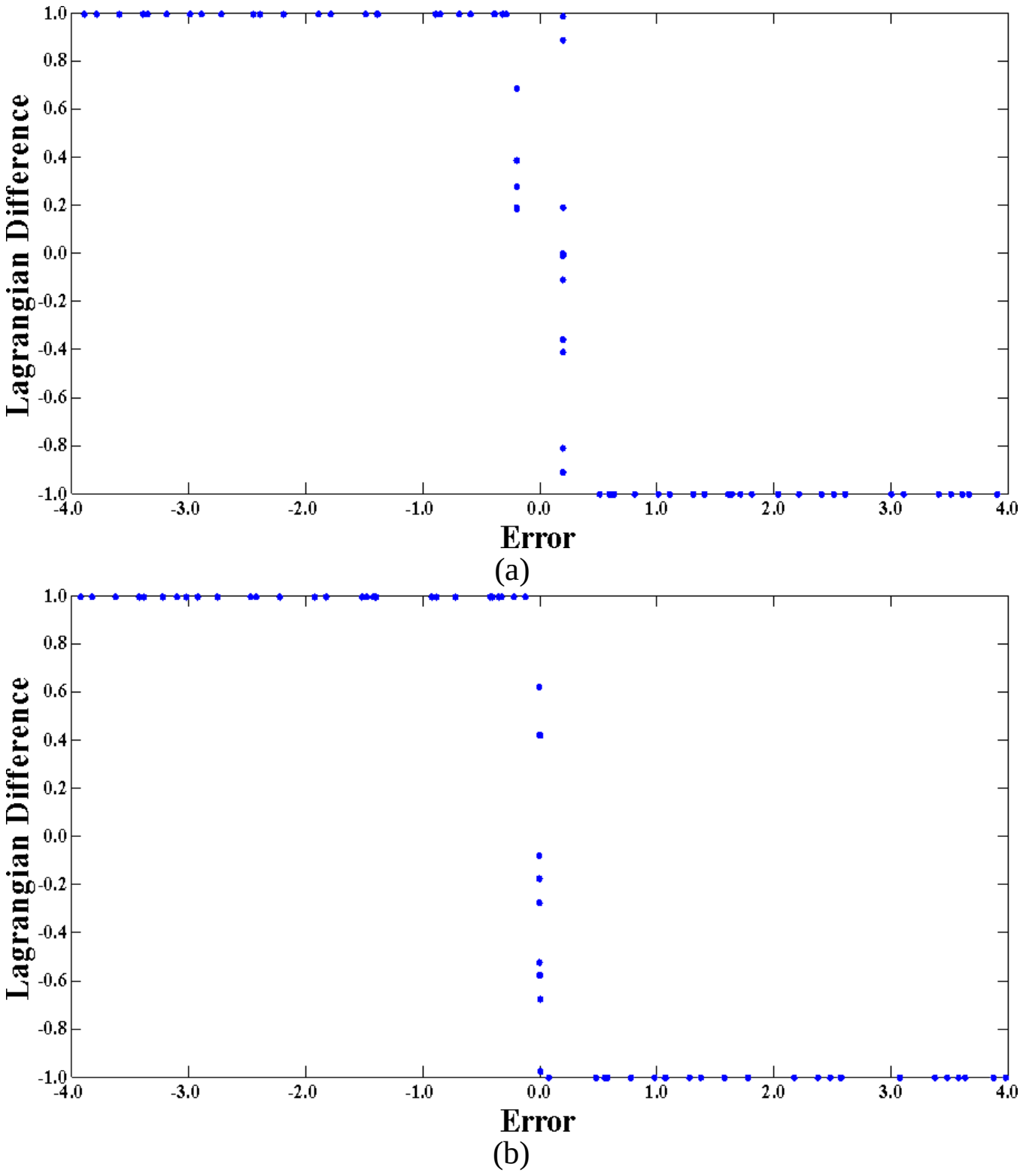

Fig. 3. WEC of typical SVR, where $\rho$ = 0.0, $C$ = 1.0, and $\varepsilon$ = 0.2 (upper part) / 0.0 (lower part). The horizontal axis specifies output errors, whereas the vertical denotes $\theta$. The distance between the upper and lower bounds is 2$C$.

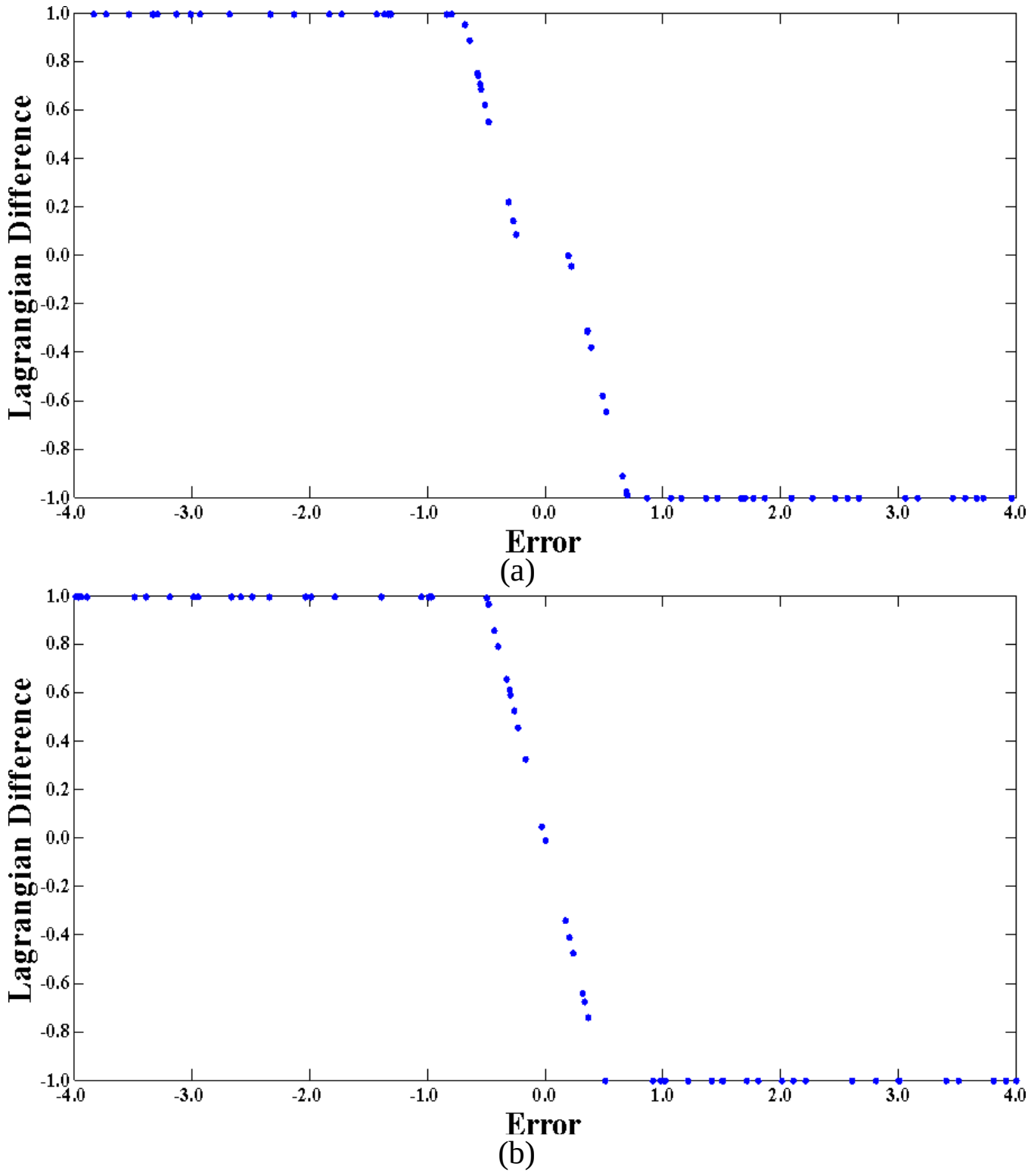

Fig. 4. WEC of Ridge SVR, where $\rho$ = 0.5, $C$ = 1.0, and $\varepsilon$ = 0.2 (upper part) / 0.0 (lower part). The horizontal axis specifies output errors, whereas the vertical denotes $\theta$. The distance between the upper and lower bounds is 2$C$.

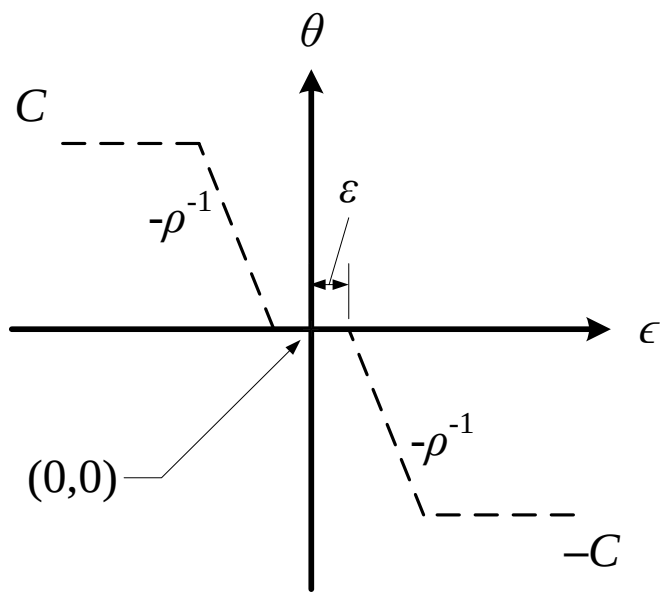

Fig. 5. Illustration of WECs. The dashed line is the WEC of Ridge SVR. The slope is $-\rho^{-1}$, and the upper/lower bounds are respectively $C$ and $-C$. When two dashed lines intersect with the axis with $\theta$=0, the distance between two intersections is 2$\varepsilon$.

Let $\theta$ denote α - α*. Subsequently,

$$\Gamma\left(\mathbf{x}_i\right)=\begin{cases}\theta_i = \rho^{-1}\varepsilon-\rho^{-1}\left(f\left(\mathbf{x}_i\right)-y_i\right), & \text{if } \epsilon_i > \varepsilon\\ \theta_i = -\rho^{-1}\varepsilon-\rho^{-1}\left(f\left(\mathbf{x}_i\right)-y_i\right), & \text{if } \epsilon_i < -\varepsilon\end{cases} \quad (36)$$

$$\text{s.t.}\begin{cases}\theta_i = C, & \text{if } \theta_i > C\\ \theta_i = -C, & \text{if } \theta_i < -C\\ \theta_i = 0, & \text{if } -\varepsilon \le \epsilon_i \le \varepsilon\end{cases}.$$

Taking the partial derivative of (35) with respect to α and α* [8] followed by calculating the difference yields

$$\frac{\partial \mathfrak{H}_{\text{SVR}}}{\partial \alpha_i} - \frac{\partial \mathfrak{H}_{\text{SVR}}}{\partial \alpha_i^*} = \sum_{j=1}^{N} \mathcal{Q}_{ij}\left(\alpha_j - \alpha_j^*\right) - y_i + b \\ = \sum_{j=1}^{N} \mathcal{Q}_{ij}\theta_j - y_i + b. \tag{37}$$

Additionally,

$$\epsilon_i = f\left(\mathbf{x}_i\right) - y_i = \sum_{j=1}^{N} \mathcal{Q}_{ij}\theta_j - y_i + b \tag{38}$$

where

$$\begin{cases} \epsilon_i \ge \varepsilon, & \theta_i = -C \\ \epsilon_i = \varepsilon, & -C < \theta_i < 0 \\ -\varepsilon \le \epsilon_i \le \varepsilon, & \theta_i = 0 \\ \epsilon_i = -\varepsilon, & 0 < \theta_i < C \\ \epsilon_i \le -\varepsilon, & \theta_i = C \end{cases}.$$

Like (11), (38) can be rewritten based on three regions $\mathcal{S}$, $\mathcal{I}$, and $\mathcal{O}$. Accordingly,

$$\mathfrak{F}_i = \sum_{j=1}^{N} \mathcal{Q}_{ij}\theta_j - y_i + b \\ = \sum_{s\in\mathcal{S}} \mathcal{Q}_{is}\theta_s + \sum_{\iota\in\mathcal{I}} \mathcal{Q}_{i\iota}\theta_\iota + \sum_{o\in\mathcal{O}} \mathcal{Q}_{io}\theta_o - y_i + b \\ = \sum_{s\in\mathcal{S}} \mathcal{Q}_{is}\theta_s + \sum_{\iota\in\mathcal{I}} \mathcal{Q}_{i\iota}\theta_\iota - y_i + b \tag{39}$$

where

$$\begin{cases} \|\mathfrak{F}_i\| - \varepsilon = 0, & 0 < \|\theta_i\| < C, & i \in \mathcal{S} \\ \|\mathfrak{F}_i\| - \varepsilon > 0, & \|\theta_i\| = C, & i \in \mathcal{I} \\ \|\mathfrak{F}_i\| - \varepsilon < 0, & \|\theta_i\| = 0, & i \in \mathcal{O} \end{cases}$$

and $\|\cdot\|$ denotes the absolute value.

Recall that $\mathcal{D}$ represents the set of new training samples, and $\mathcal{R}$ denotes the set of existing training samples that are about to be deleted. The multiple incremental and decremental equation is

$$\Delta\mathfrak{F}_i = \sum_{d\in\mathcal{D}} \mathcal{Q}_{id}\Delta\theta_d + \sum_{r\in\mathcal{R}} \mathcal{Q}_{ir}\Delta\theta_r + \sum_{s\in\mathcal{S}} \mathcal{Q}_{is}\Delta\theta_s + \Delta b \\ = 0,\ i \in \mathcal{S} \tag{40}$$

and

$$\mathfrak{O} = \sum_{d\in\mathcal{D}} \Delta\theta_d + \sum_{r\in\mathcal{R}} \Delta\theta_r + \sum_{s\in\mathcal{S}} \Delta\theta_s = 0\,. \tag{41}$$

Any change in the existing training data should maintain the equilibrium modeled by (40) and (41). Equations (40) and (41) can be converted into a matrix form as follows.

$$\begin{bmatrix} 0 & \mathbf{1}_{\mathcal{S}}^{\mathrm{T}} \\ \mathbf{1}_{\mathcal{S}} & \boldsymbol{\mathcal{Q}}_{\mathcal{S}} \end{bmatrix} \begin{bmatrix} \Delta b \\ \Delta\boldsymbol{\theta}_{\mathcal{S}} \end{bmatrix} + \begin{bmatrix} \mathbf{1}_{\mathcal{D}}^{\mathrm{T}} & \mathbf{1}_{\mathcal{R}}^{\mathrm{T}} \\ \boldsymbol{\mathcal{Q}}_{\mathcal{S},\mathcal{D}} & \boldsymbol{\mathcal{Q}}_{\mathcal{S},\mathcal{R}} \end{bmatrix} \begin{bmatrix} \Delta\boldsymbol{\theta}_{\mathcal{D}} \\ \Delta\boldsymbol{\theta}_{\mathcal{R}} \end{bmatrix} = 0\,. \tag{42}$$

When gradual step sizes $\eta'$ are applied, then

$$\Delta\boldsymbol{\theta}_{\mathcal{D}} = \eta'\left(-\mathrm{sgn}_{\mathcal{D}} \otimes \left(C\mathbf{1}\right) - \boldsymbol{\theta}_{\mathcal{D}}\right). \tag{43}$$

and

$$\Delta\boldsymbol{\theta}_{\mathcal{R}} = \eta'\left(\mathrm{sgn}_{\mathcal{R}} \otimes \boldsymbol{\theta}_{\mathcal{R}}\right) \tag{44}$$

where sgn(·) returns the sign of $\epsilon_i$-$\varepsilon$. Based on (39) and (40),

$$\begin{cases} \|\mathfrak{F}_i\| = \varepsilon \\ \Delta\mathfrak{F}_i = 0 \end{cases} \Leftrightarrow \|\mathfrak{F}_i\| + \Delta\mathfrak{F}_i = \varepsilon, \quad i \in \mathcal{S}\,. \tag{45}$$

Like (23) and (25), there is an equation for the maximum change $\boldsymbol{\varphi}'$ (based on $\mathcal{D}$ and $\mathcal{R}$) and $\mathcal{S}$ Thus,

$$\Delta\mathfrak{F}_i = \varepsilon - \|\mathfrak{F}_i\| = \eta'\varphi_i'\,. \tag{46}$$

Subsequently calculating the minimum amount of step sizes by checking all the training samples plus $\mathcal{D}$ and $\mathcal{R}$ yields

$$\eta'^* = \arg\min_{\eta'} \left\| \left(\varepsilon - \|\mathfrak{F}_i\|\right) / \varphi_i' \right\|. \tag{47}$$

When the mechanism of Ridge SVR is applied, let

$$\Delta\boldsymbol{\theta}_{\mathcal{D}} = \omega\left(\mathbf{x}_{\mathcal{D}}\right). \tag{48}$$

and

$$\Delta\boldsymbol{\theta}_{\mathcal{R}} = \mathrm{sgn}_{\mathcal{R}} \otimes \boldsymbol{\theta}_{\mathcal{R}} \tag{49}$$

where ω(·) is the combination of (36).

To avoid generating inappropriate $\theta$, the system rechecks (39) after $\theta$ is predicted by using (48) and (49). Any values that violate (39) are adjusted to fit the conditions. Thus,

$$\begin{aligned}\begin{bmatrix}\Delta b\\ \Delta\boldsymbol{\theta}_{\mathcal{S}}\end{bmatrix} &= -\begin{bmatrix}0 & \mathbf{1}_{\mathcal{S}}^{\mathrm{T}}\\ \mathbf{1}_{\mathcal{S}} & \mathcal{Q}_{\mathcal{S}}\end{bmatrix}^{-1}\begin{bmatrix}\mathbf{1}_{\mathcal{D}}^{\mathrm{T}} & \mathbf{1}_{\mathcal{R}}^{\mathrm{T}}\\ \mathcal{Q}_{\mathcal{S},\mathcal{D}} & \mathcal{Q}_{\mathcal{S},\mathcal{R}}\end{bmatrix}\begin{bmatrix}\Delta\boldsymbol{\theta}_{\mathcal{D}}\\ \Delta\boldsymbol{\theta}_{\mathcal{R}}\end{bmatrix}\\ &= -\begin{bmatrix}0 & \mathbf{1}_{\mathcal{S}}^{\mathrm{T}}\\ \mathbf{1}_{\mathcal{S}} & \mathcal{Q}_{\mathcal{S}}\end{bmatrix}^{-1}\begin{bmatrix}\mathbf{1}_{\mathcal{D}}^{\mathrm{T}} & \mathbf{1}_{\mathcal{R}}^{\mathrm{T}}\\ \mathcal{Q}_{\mathcal{S},\mathcal{D}} & \mathcal{Q}_{\mathcal{S},\mathcal{R}}\end{bmatrix}\begin{bmatrix}\omega(\mathbf{x}_{\mathcal{D}})\\ \mathrm{sgn}_{\mathcal{R}}\otimes\boldsymbol{\theta}_{\mathcal{R}}\end{bmatrix}\end{aligned}. \quad (50)$$

Based on the Schur complement theory,

$$\begin{bmatrix}0 & \mathbf{1}_{\mathcal{S}}^{\mathrm{T}}\\ \mathbf{1}_{\mathcal{S}} & \mathcal{Q}_{\mathcal{S}}\end{bmatrix}^{-1} = \begin{bmatrix}z' & -z'\mathbf{1}_{\mathcal{S}}^{\mathrm{T}}\mathcal{Q}_{\mathcal{S}}^{-1}\\ -z'\mathcal{Q}_{\mathcal{S}}^{-1}\mathbf{1}_{\mathcal{S}} & z'\mathcal{Q}_{\mathcal{S}}^{-1}\mathbf{1}_{\mathcal{S}}\mathbf{1}_{\mathcal{S}}^{\mathrm{T}}\mathcal{Q}_{\mathcal{S}}^{-1} + \mathcal{Q}_{\mathcal{S}}^{-1}\end{bmatrix} \quad (51)$$

where $z'$ is a scalar derived from the inverse of $\mathcal{Q}_{\mathcal{S}}$ shown as follows:

$$z' = -\left(\mathbf{1}_{\mathcal{S}}^{\mathrm{T}}\mathcal{Q}_{\mathcal{S}}^{-1}\mathbf{1}_{\mathcal{S}}\right)^{-1}. \quad (52)$$

For empty $\mathcal{S}$, the system is required to select two new training samples to rebuild $\mathcal{S}$ based on a closed-form equation in Martin [8] and Ma *et al*. [9].

## VI. Experimental Results

Experiments on four open datasets were carried out for evaluating the performance. The information of these datasets is listed in Table II. The first column shows the name. The rest columns specify the number of classes, samples, and dimensions, respectively. For classification, dataset "MIT/BIH ECG" is available at PhysioNet (www.physionet.org), and "Skin Segmentation (SS)" is from the UC Irvine (UCI) Machine Learning Repository (archive.ics.uci.edu/ml/). Regarding regression, datasets "Combined Cycle Power Plants (CCPP)" and "Household Power Consumption (HPC)" were downloaded from the UCI Machine Learning Repository.

The experiment used approximately 80% of the data for training and 10% of the data for multiple incremental training. The rest 10% were for testing. Furthermore, +6/-2 and +40/-10 samples were randomly selected for multiple incremental and decremental learning at the same time. Table III summarizes the incremental and decremental settings. For algorithms, three typical kernels were used herein — Second-ordered polynomial (poly2), third-ordered polynomial (poly3), and radial basis functions (with the empirical deviation equal to 50 for classification and 2/4 for regression). All the ridge parameters were 0.5. For fairness, we selected the system developed by Karasuyama and Takeuchi [6] as the baseline because both their system and our method supported multiple incremental and decremental analyses. Before training, all the samples were standardized. The labels, only for regression, were standardized.

Table II
Attributes of the Datasets for Classification & Regression

| Name | #Classes | #Samples | #Dimensions |
|---|---|---|---|
| ECG | 2 | 26008 | 21 |
| SS | 2 | 24505 | 3 |
| HPC | N/A | 20378 | 7 |
| CCPP | N/A | 26910 | 4 |

Table III
Settings of Multiple Incremental/Decremental Learning

| Name | Basic Training Size | Multiple Incremental/Decremental Size |
|---|---|---|
| ECG | 20806 | +6 / -2 |
| SS | 19604 | +40 / -10 |
| HPC | 16302 | +6 / -2 |
| CCPP | 21528 | +40 / -10 |

Table IV
Algorithmic Settings

| Method | Kernel | Ridge | RBF Sigma |
|---|---|---|---|
| Karasuyama & Takeuchi [5] | Poly2, Poly3, & RBFs | N/A | 50 (Classification)<br>4 (Regression) |
| Proposed | Poly2, Poly3, & RBFs | 0.5 | 50 (Classification)<br>2 & 4 (Regression) |

*Deviation of RBFs was empirically set, and the penalty was one.

As the system developed by Karasuyama and Takeuchi [6] and the proposed algorithms generate the same optimal solutions as the original nonincremental algorithm does, performance in accuracy remains unchanged. Thus, the metric for evaluating multiple incremental/decremental algorithms is computational time.

### A. *Multiple Incremental/Decremental Ridge SVMs*

Fig. 6–Fig. 11 display the multiple incremental results, where the horizontal axis denotes the round, and the vertical axis represents the accumulative computational time in log10. The solid line indicates the result generated by the proposed method, whereas the dashed line signifies the approach by Karasuyama and Takeuchi [6]. Nonincremental results are displayed by using curves with plus signs.

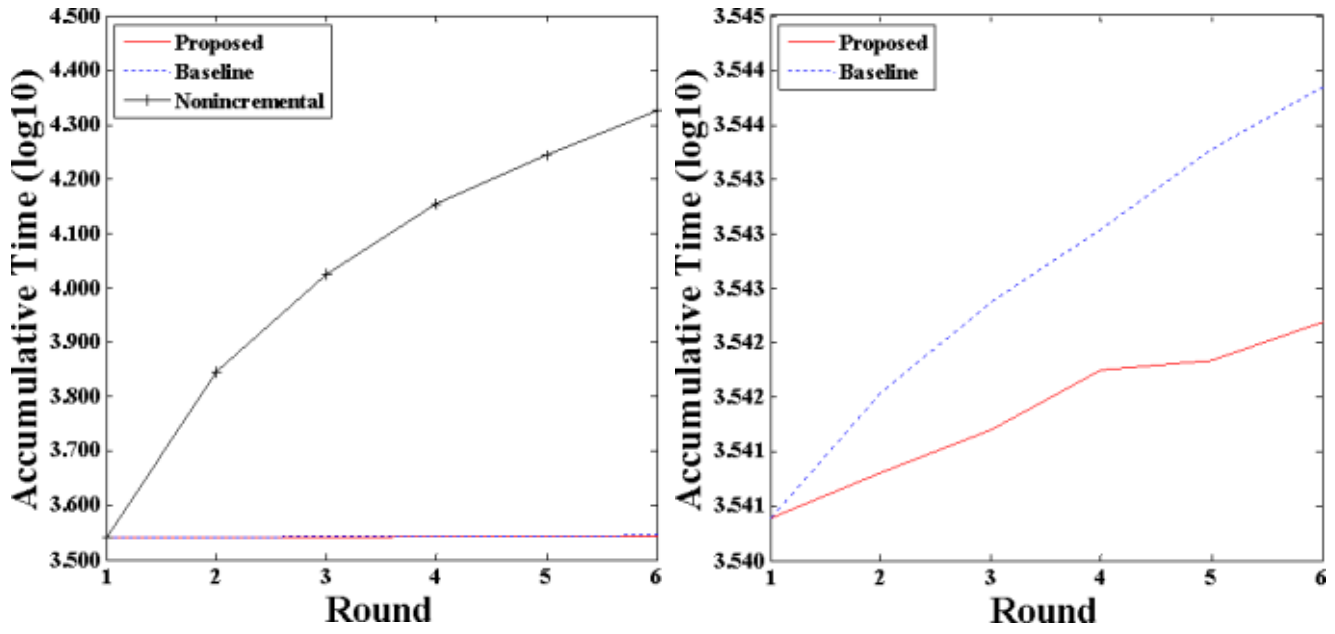

Fig. 6. Comparison between the proposed (red), the baseline (blue), and the nonincremental (black) learning with the use of the ECG dataset and the poly2 kernel. The accuracy rates are 95.10%. Left: Three methods. Right: A closer view of the two methods

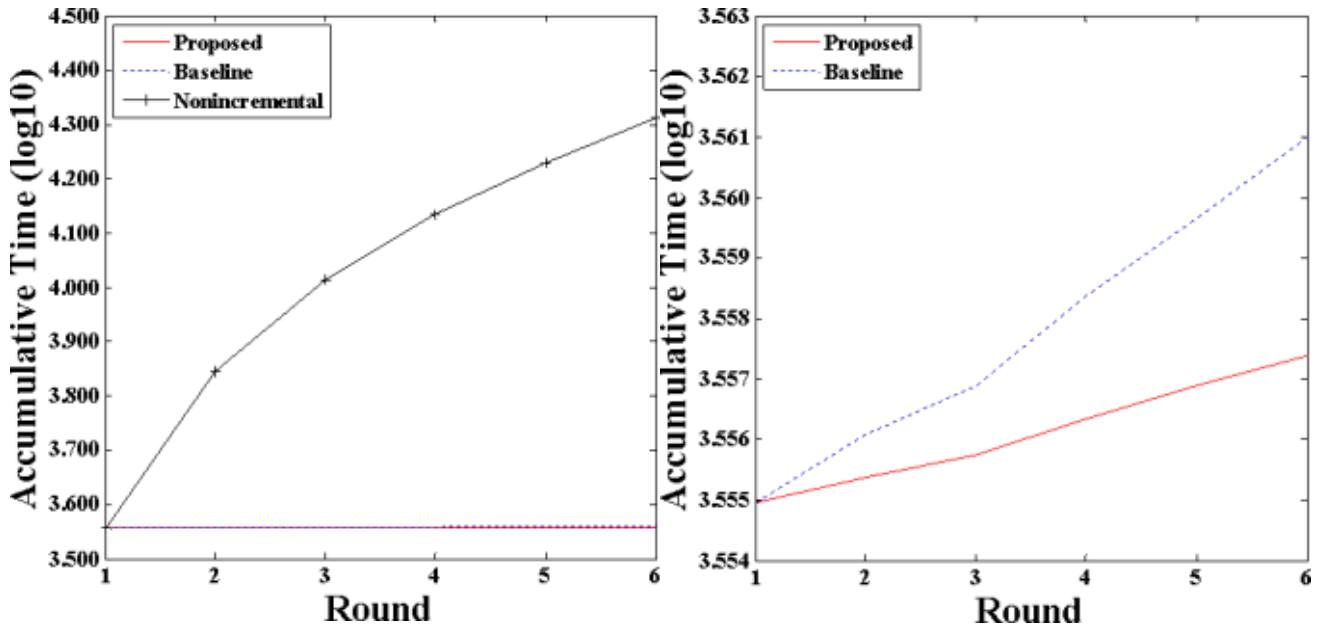

Fig. 7. Comparison between the proposed (red), the baseline (blue), and the nonincremental (black) learning with the use of the ECG dataset and the poly3 kernel. The accuracy rates are 97.81%. Left: Three methods. Right: A closer view of the two methods

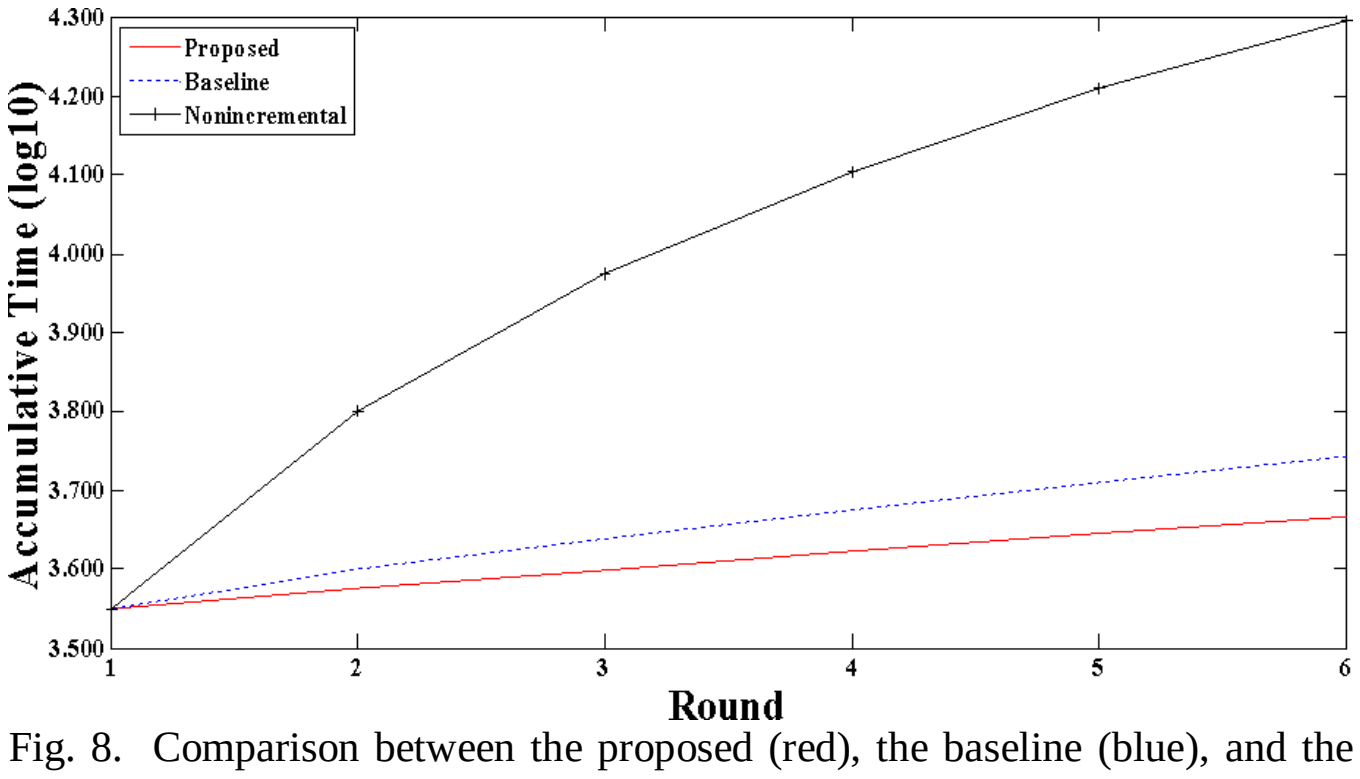

Fig. 8. Comparison between the proposed (red), the baseline (blue), and the nonincremental (black) learning with the use of the ECG dataset and the RBF. The accuracy rates are 97.40%.

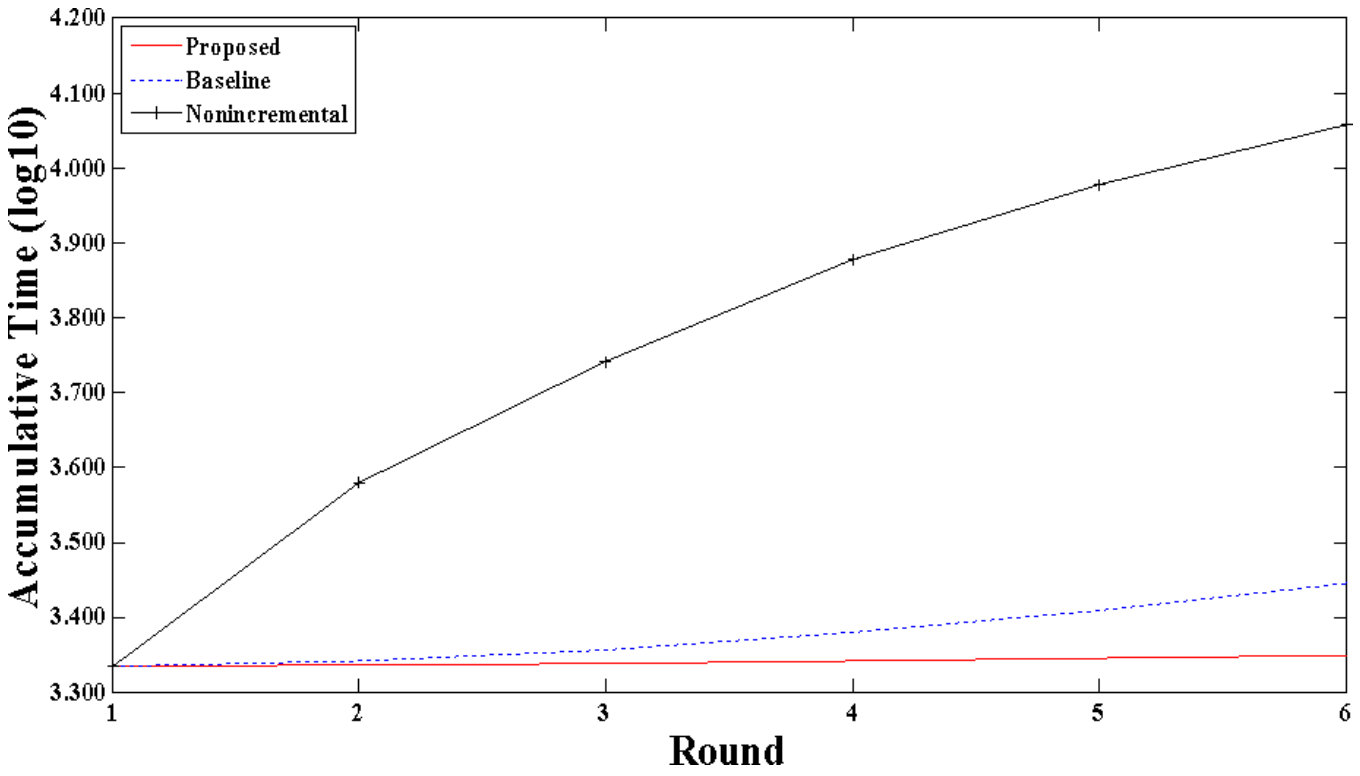

Fig. 9. Comparison between the proposed (red), the baseline (blue), and the nonincremental (black) learning with the use of the SS dataset and the poly2 kernel. The accuracy rates are 99.80%.

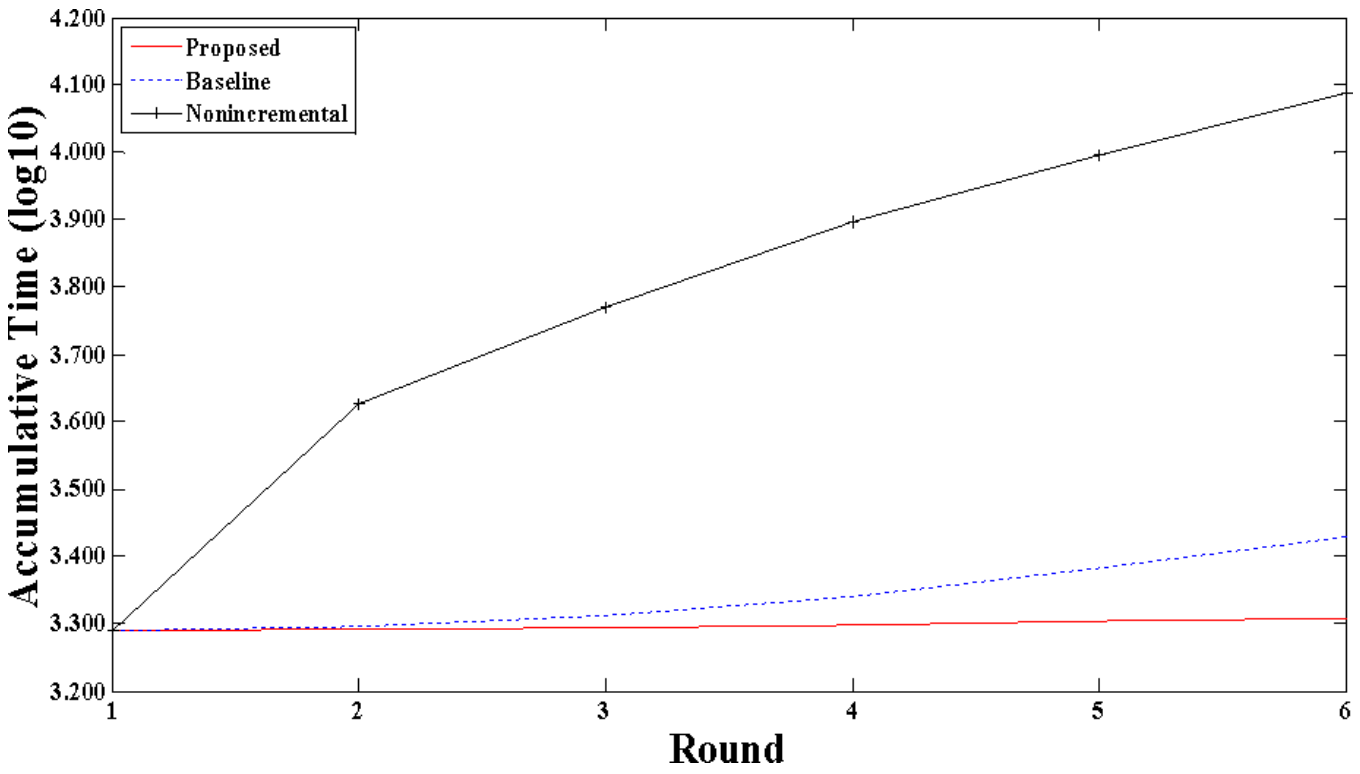

Fig. 10. Comparison between the proposed (red), the baseline (blue), and the nonincremental (black) learning with the use of the SS dataset and the poly3 kernel. The accuracy rates are 94.96%.

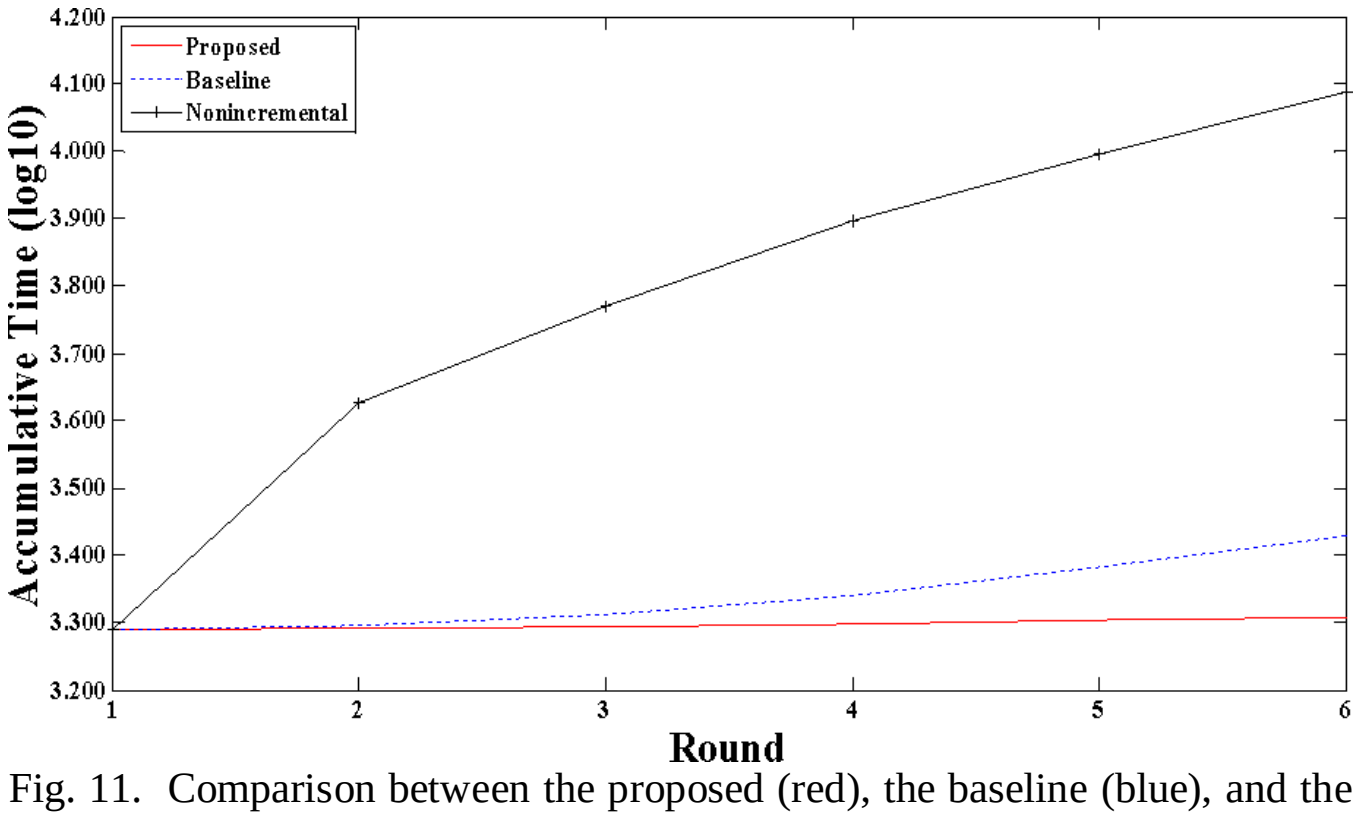

Fig. 11. Comparison between the proposed (red), the baseline (blue), and the nonincremental (black) learning with the use of the SS dataset and the RBF. The accuracy rates are 99.84%.

Table V
Computational Time Based on the ECG Dataset and the Poly2 Kernel

| #Samples | 20810 | 20814 | 20818 | 20822 | 20826 |
|---|---|---|---|---|---|
| Proposed | 3.34 | 3.15 | 4.39 | 0.73 | 2.74 |
| Baseline [6] | 9.26 | 6.62 | 5.36 | 5.90 | 4.58 |
| Nonincremental | 3540.99 | 3598.53 | 3630.37 | 3309.53 | 3634.98 |

Unit is seconds

Table VI
Computational Time Based on the ECG Dataset and the Poly3 Kernel

| #Samples | 20810 | 20814 | 20818 | 20822 | 20826 |
|---|---|---|---|---|---|
| Proposed | 3.31 | 3.17 | 4.88 | 4.63 | 4.10 |
| Baseline [6] | 9.19 | 6.86 | 12.27 | 10.65 | 11.31 |
| Nonincremental | 3399.23 | 3339.26 | 3282.82 | 3378.43 | 3528.98 |

Unit is seconds

Table VII
Computational Time Based on the ECG Dataset and the RBF

| #Samples | 20810 | 20814 | 20818 | 20822 | 20826 |
|---|---|---|---|---|---|
| Proposed | 218.60 | 215.18 | 226.34 | 222.87 | 213.33 |
| Baseline [6] | 434.57 | 378.36 | 374.86 | 401.76 | 406.36 |
| Nonincremental | 2770.66 | 3111.16 | 3287.58 | 3546.36 | 3489.52 |

Unit is seconds

Table VIII
Computational Time Based on the SS Dataset and the Poly2 Kernel

| #Samples | 19634 | 19664 | 19694 | 19724 | 19754 |
|---|---|---|---|---|---|
| Proposed | 5.75 | 14.95 | 11.66 | 24.83 | 18.28 |
| Baseline [6] | 33.74 | 76.75 | 125.56 | 170.59 | 216.31 |
| Nonincremental | 1637.61 | 1714.72 | 2018.80 | 1944.05 | 1923.78 |

Unit is seconds

Table IX
Computational Time Based on the SS Dataset and the Poly3 Kernel

| #Samples | 19634 | 19664 | 19694 | 19724 | 19754 |
|---|---|---|---|---|---|
| Proposed | 7.25 | 9.73 | 19.85 | 21.85 | 22.68 |
| Baseline [6] | 27.91 | 71.26 | 142.25 | 215.18 | 279.31 |
| Nonincremental | 2273.97 | 1644.35 | 2017.36 | 1991.27 | 2384.43 |

Unit is seconds

Table X
Computational Time Based on the SS Dataset and the RBF

| #Samples | 19634 | 19664 | 19694 | 19724 | 19754 |
|---|---|---|---|---|---|
| Proposed | 8.75 | 3.37 | 2.95 | 4.04 | 3.05 |
| Baseline [6] | 445.39 | 458.62 | 476.55 | 498.43 | 514.96 |
| Nonincremental | 2045.49 | 1795.20 | 1832.76 | 2086.76 | 2102.43 |

Unit is seconds

Table XI
Average Improvement in Computational Time

| | Proposed | Baseline [6] | Enhancement |
|---|---|---|---|
| ECG — Poly2 | 2.87 | 6.35 | 120.77% |
| ECG — Poly3 | 4.02 | 10.05 | 150.06% |
| ECG — RBF | 219.26 | 399.18 | 82.05% |
| SS — Poly2 | 15.09 | 124.59 | 725.18% |
| SS — Poly3 | 16.27 | 147.18 | 804.18% |
| SS — RBF | 4.43 | 478.79 | 10690.63% |

Unit is seconds

Examining these figures reveals that incremental and decremental mechanisms indeed saved the computational load. Table V–Table XI summarize the numeric results of the experiments. The findings showed that the incremental/ decremental mechanism for SVMs could improve the efficiency by 20 times on average compared with the baseline.

## B. *Multiple Incremental/Decremental Ridge SVR*

This subsection evaluates the performance of the proposed method and the baseline. The experimental settings are all described at the beginning of Section VI. For performance indicators, this experiment measured computational time and mean squared errors (MSEs).

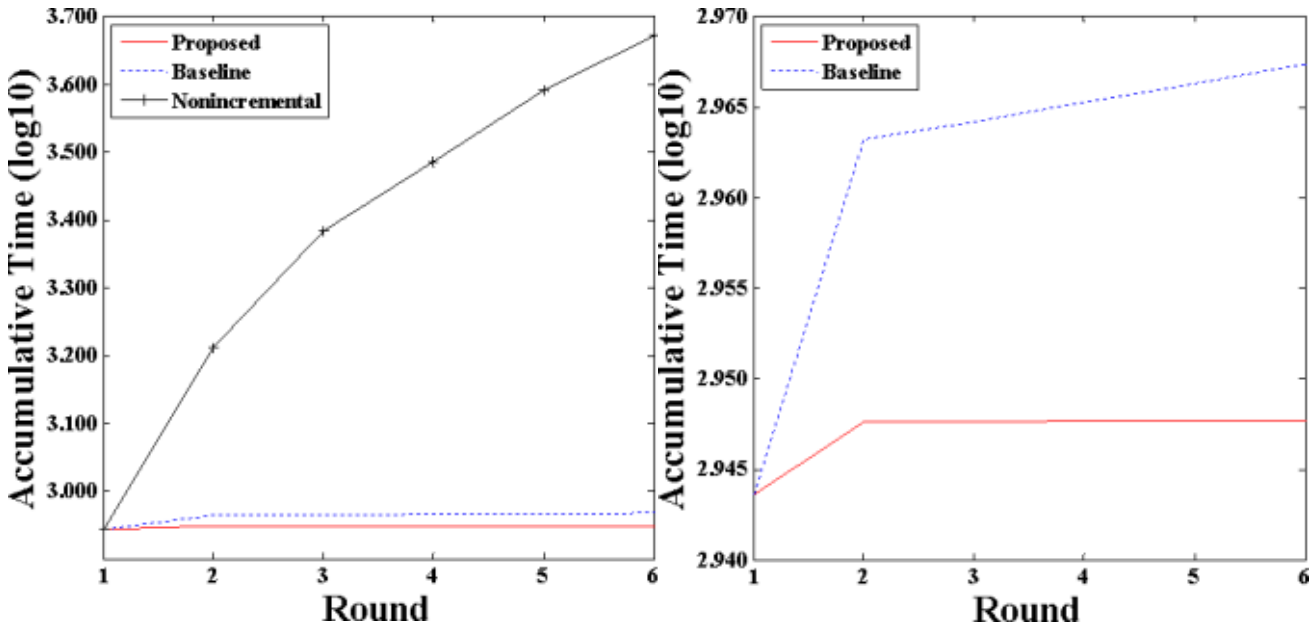

Fig. 12. Comparison between the proposed (red), the baseline (blue), and the nonincremental (black) learning with the use of the HPC dataset and the poly2 kernel. The MSE is 0.0646. Left: Three methods. Right: A closer view of the two methods

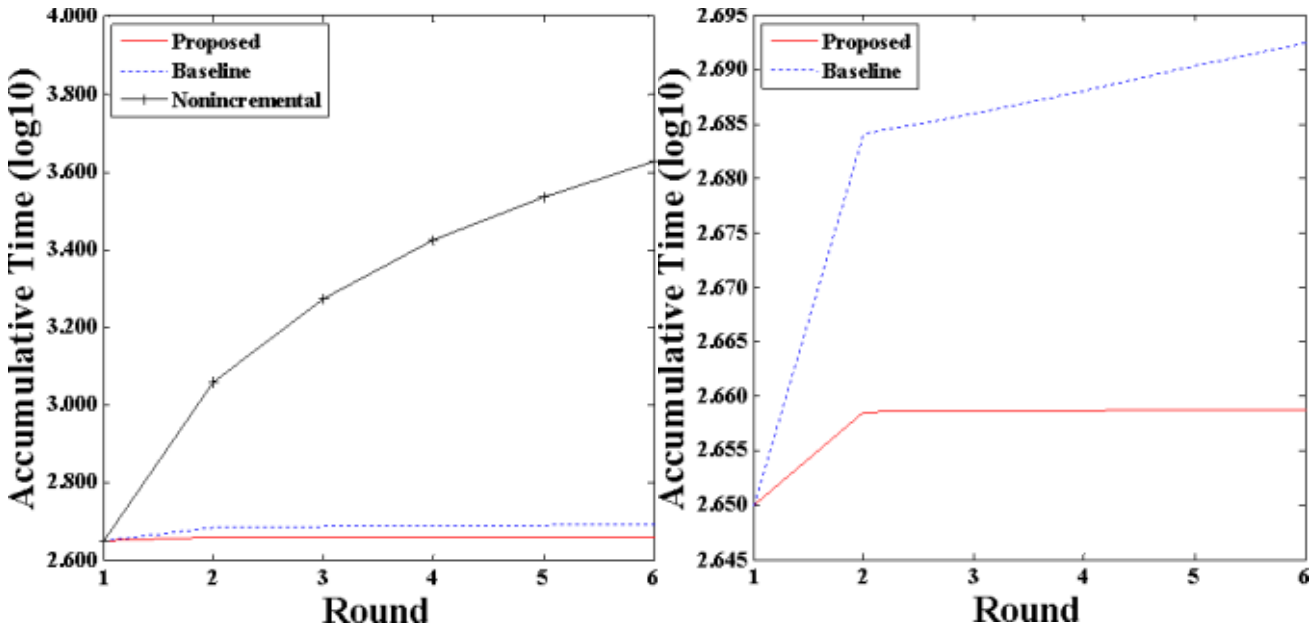

Fig. 13. Comparison between the proposed (red), the baseline (blue), and the nonincremental (black) learning with the use of the HPC dataset and the poly3 kernel. The MSE is 0.0517. Left: Three methods. Right: A closer view of the two methods

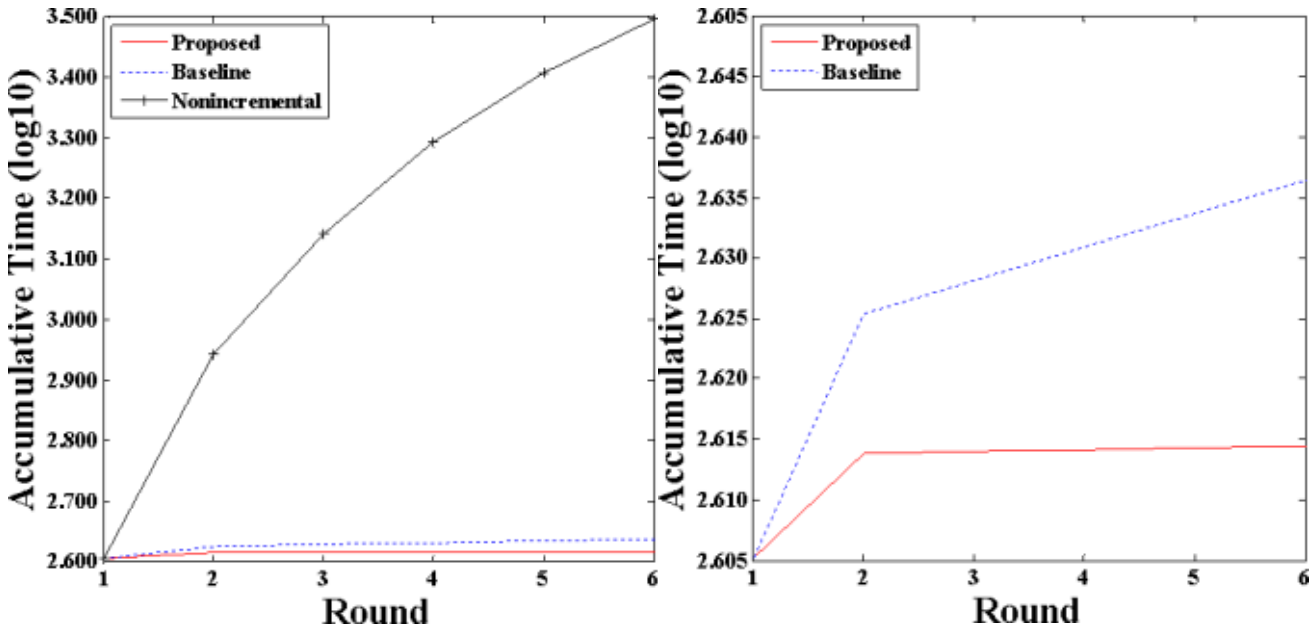

Fig. 14. Comparison between the proposed (red), the baseline (blue), and the nonincremental (black) learning with the use of the HPC dataset and the RBF. The MSE is 0.0641. Left: Three methods. Right: A closer view of the two methods

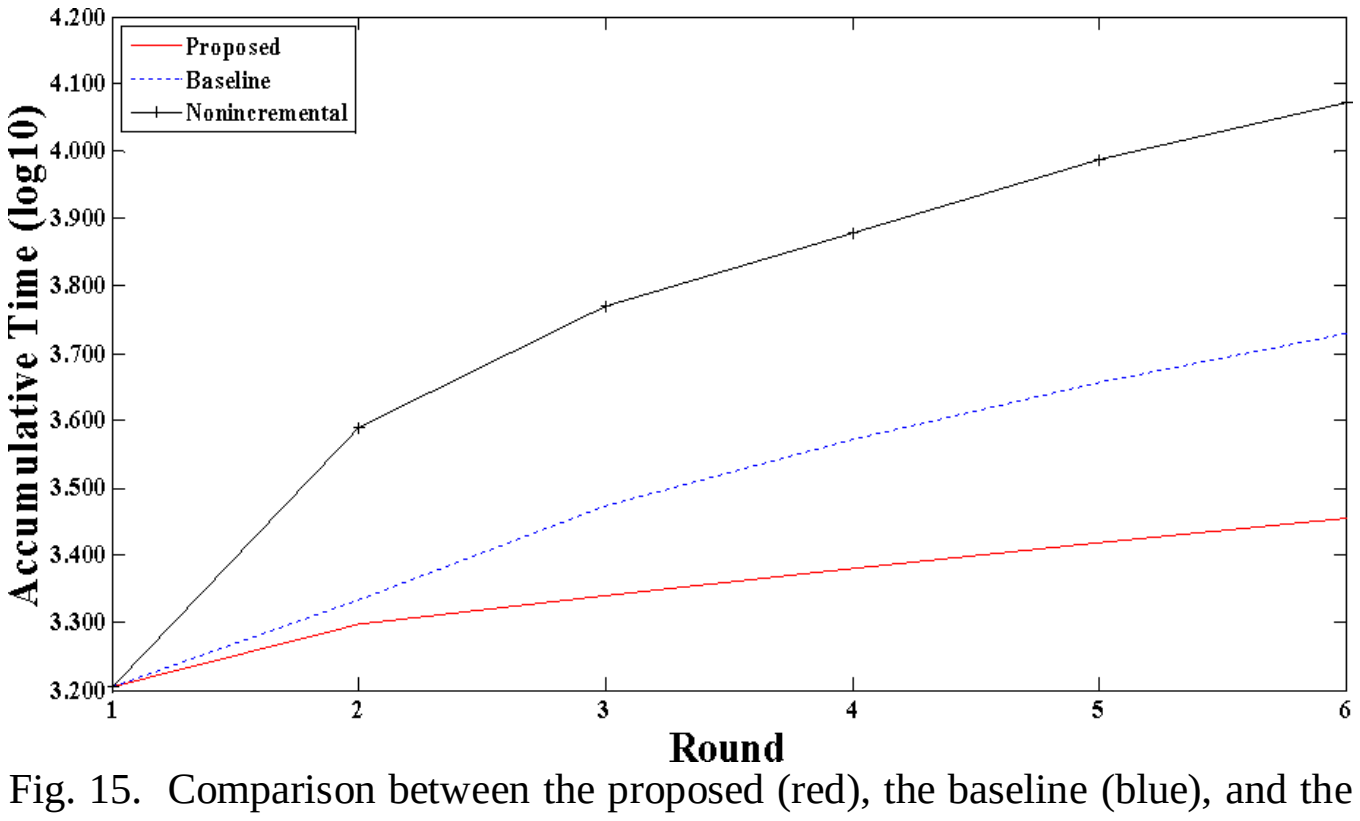

Fig. 15. Comparison between the proposed (red), the baseline (blue), and the nonincremental (black) learning with the use of the CCPP dataset and the poly2 kernel. The MSE is 0.0104.

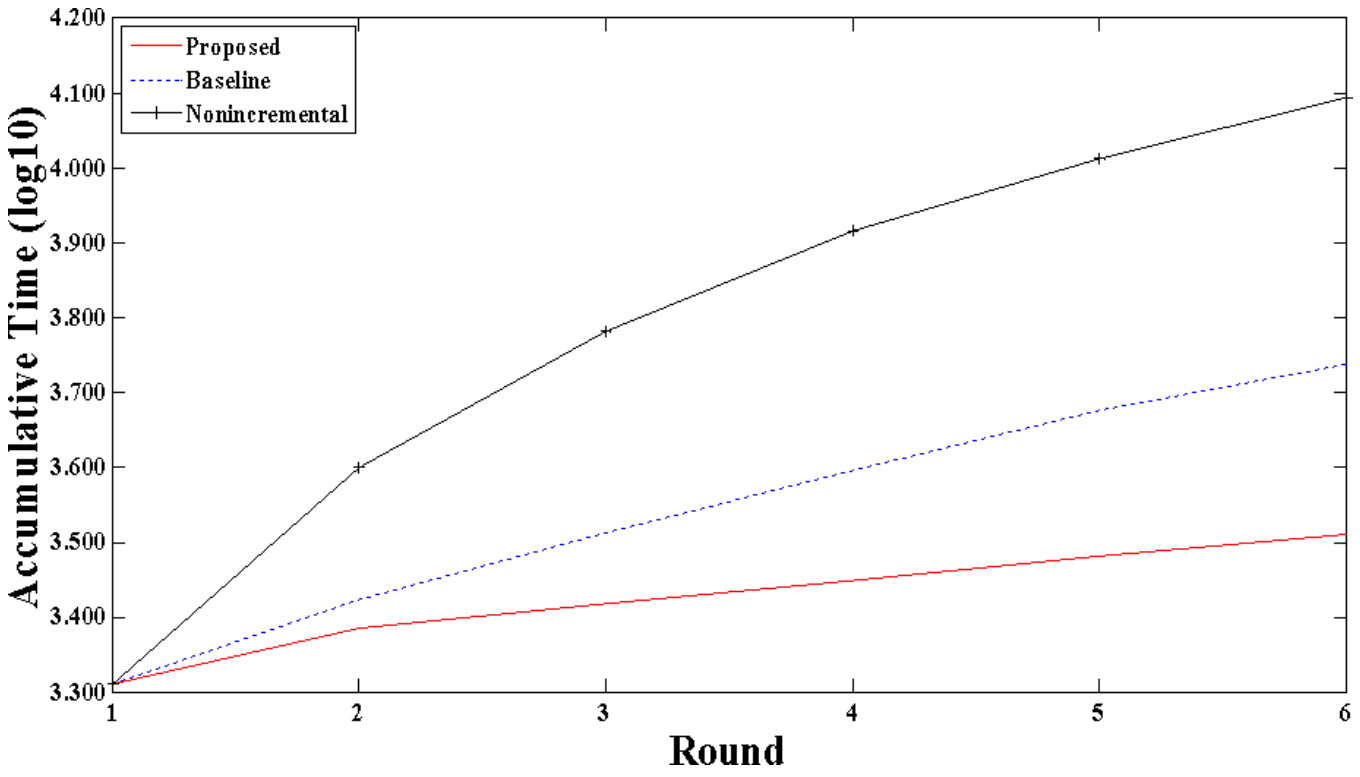

Fig. 16. Comparison between the proposed (red), the baseline (blue), and the nonincremental (black) learning with the use of the CCPP dataset and the poly3 kernel. The MSE is 0.0097.

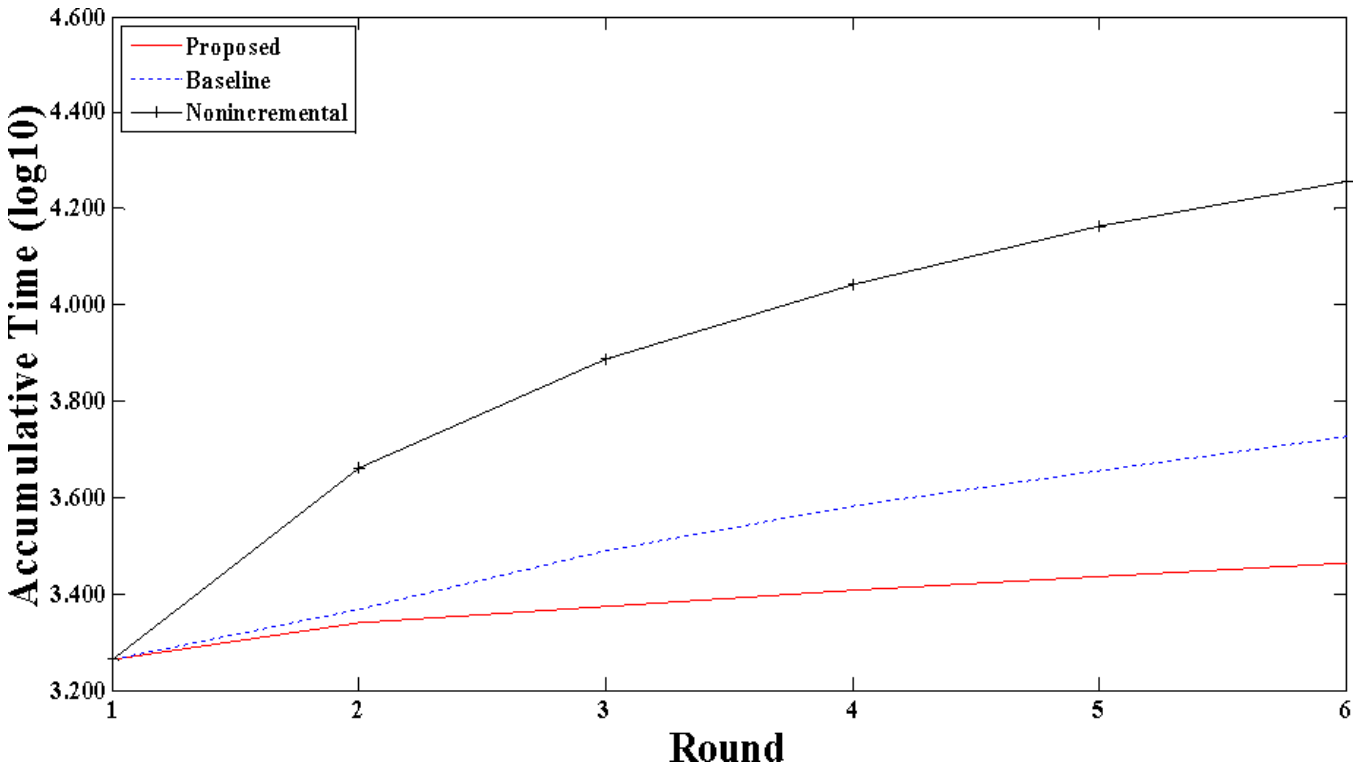

Fig. 17. Comparison between the proposed (red), the baseline (blue), and the nonincremental (black) learning with the use of the CCPP dataset and the RBF. The MSE is 0.100.

Table XII
Computational Time Based on the HPC Dataset and the Poly2 Kernel

| #Samples | 16306 | 16310 | 16314 | 16318 | 16322 |
|---|---|---|---|---|---|
| Proposed | 8.04 | 0.07 | 0.05 | 0.02 | 0.02 |
| Baseline [6] | 40.54 | 2.00 | 2.30 | 2.24 | 2.22 |
| Nonincremental | 746.26 | 791.20 | 636.18 | 845.24 | 794.35 |

Unit is seconds

Table XIII
Computational Time Based on the HPC Dataset and the Poly3 Kernel

| #Samples | 16306 | 16310 | 16314 | 16318 | 16322 |
|---|---|---|---|---|---|
| Proposed | 8.89 | 0.05 | 000.08 | 000.03 | 000.03 |
| Baseline [6] | 36.49 | 2.11 | 002.34 | 002.58 | 002.27 |
| Nonincremental | 701.50 | 726.32 | 780.62 | 783.34 | 790.35 |

Unit is seconds

Table XIV
Computational Time Based on the HPC Dataset and the RBF

| #Samples | 16306 | 16310 | 16314 | 16318 | 16322 |
|---|---|---|---|---|---|
| Proposed | 8.11 | 0.13 | 0.14 | 000.13 | 000.14 |
| Baseline [6] | 19.15 | 2.68 | 2.72 | 002.76 | 002.68 |
| Nonincremental | 473.69 | 509.46 | 579.78 | 590.23 | 587.33 |

Unit is seconds

Table XV
Computational Time Based on the CCPP Dataset and the Poly2 Kernel

| #Samples | 21558 | 21588 | 21618 | 21648 | 21678 |
|---|---|---|---|---|---|
| Proposed | 379.37 | 203.50 | 215.99 | 218.24 | 224.00 |
| Baseline [6] | 554.01 | 814.54 | 753.53 | 811.69 | 814.32 |
| Nonincremental | 2283.47 | 1983.90 | 1691.83 | 2158.53 | 2100.32 |

Unit is seconds

Table XVI
Computational Time Based on the CCPP Dataset and the Poly3 Kernel

| #Samples | 21558 | 21588 | 21618 | 21648 | 21678 |
|---|---|---|---|---|---|
| Proposed | 386.61 | 183.73 | 196.20 | 218.75 | 213.59 |
| Baseline [6] | 613.24 | 596.58 | 700.42 | 785.62 | 732.94 |
| Nonincremental | 1934.01 | 2055.62 | 2188.29 | 2042.4 | 2162.43 |

Unit is seconds

Table XVII
Computational Time Based on the CCPP Dataset and the RBF

| #Samples | 21558 | 21588 | 21618 | 21648 | 21678 |
|---|---|---|---|---|---|
| Proposed | 353.91 | 178.09 | 188.10 | 179.79 | 176.42 |
| Baseline [6] | 508.66 | 745.39 | 0750.91 | 0699.89 | 0791.77 |
| Nonincremental | 1601.74 | 2021.36 | 1713.33 | 2035.42 | 1943.56 |

Unit is seconds

Table XVIII
Average Improvement in Computational Time

| | Proposed | Baseline [6] | Enhancement |
|---|---|---|---|
| HPC — Poly2 | 1.64 | 9.86 | 498.90% |
| HPC — Poly3 | 1.81 | 9.16 | 403.56% |
| HPC — RBF | 1.73 | 6.00 | 246.52% |
| CCPP — Poly2 | 248.22 | 749.62 | 201.99% |
| CCPP — Poly3 | 239.78 | 685.76 | 185.99% |
| CCPP — RBF | 215.26 | 699.33 | 224.86% |

Unit is seconds

The observations on Fig. 12–Fig. 17 and Table XII–Table XVIII showed that the proposed multiple incremental/decremental mechanism for SVRs could improve the efficiency by approximately 2–5 times on average compared with the baseline.

## VII. Conclusion

This work presents an efficient incremental/decremental mechanism for Ridge SVMs and Ridge SVR, where the proposed recursive-free computation is used for high-speed online learning. With two important properties derived from ridge support vector models — Ridge parameters and WEC functions, the major problems in multiple incremental/decremental support vector learning are resolved. More flexible regions for selecting the Lagrangian multipliers of support vectors are generated. Second, no recursive computation is required for computing new Lagrangian multipliers when new instances arrive. The system can predict all the new Lagrangian multipliers at once. Moreover, the system no longer relies on typical bookkeeping strategies. These all increase overall efficiency.

Open benchmark datasets were used to evaluate the computational performance. Compared with the baseline that required gradual step sizes for multiple incremental/decremental analyses, the computational speed of the proposed method was enhanced by 20 times on average for Ridge SVMs, far faster than the baseline. Regarding multiple incremental/decremental learning for Ridge SVR, the speed was approximately 2–5 times faster than that of the baseline. Such findings have established the effectiveness of the incremental/decremental analyses.

## Appendix

This appendix elaborates how to combine the computation of the matrix inverse in multiple incremental and decremental processes at the same time. When performing multiple incremental and decremental analyses, the system needs to compute the matrix inverse in (31) and (50) frequently. To save the earlier result for updating the system in the future, namely, $\mathbf{Q}^{-1}[\ell]$ and $\mathcal{Q}^{-1}[\ell]$, we can use the following technique to compute $\mathbf{Q}^{-1}[\ell+1]$ and $\mathcal{Q}^{-1}[\ell+1]$. As $\mathbf{Q}^{-1}$ and $\mathcal{Q}^{-1}$ share the same structure, the following content uses $\mathbf{Q}^{-1}$ as an example to illustrate the process.

For multiple incremental analyses, assume $\mathcal{D}$ represents the set of new training samples. Therefore,

$$\mathbf{Q}^{-1}[\ell+1]=\begin{bmatrix}\mathbf{Q}[\ell] & \mathbf{Q}_{:,\mathcal{D}}\\ \mathbf{Q}_{:,\mathcal{D}}^{\mathrm{T}} & \mathbf{Q}_{\mathcal{D},\mathcal{D}}\end{bmatrix}^{-1}. \tag{53}$$

Based on the Sherman-Morrison formula and Woodbury matrix identity [5, 16], (53) can be decomposed to two states. One is the current state $\mathbf{Q}^{-1}[\ell]$, and the other is $\mathbf{Q}^{-1}[\ell+1]$, shown as follows.

$$\begin{aligned}\mathbf{Q}^{-1}[\ell+1]&=\begin{bmatrix}\mathbf{Q}[\ell] & \mathbf{Q}_{:,\mathcal{D}}\\ \mathbf{Q}_{:,\mathcal{D}}^{\mathrm{T}} & \mathbf{Q}_{\mathcal{D},\mathcal{D}}\end{bmatrix}^{-1}\\ &=\begin{bmatrix}\mathbf{Q}^{-1}[\ell]+\mathbf{H}_{:,\mathcal{D}}\mathbf{V}^{-1}\mathbf{H}_{:,\mathcal{D}}^{\mathrm{T}} & \mathbf{H}_{:,\mathcal{D}}\mathbf{V}^{-\mathrm{T}}\\ \mathbf{V}^{-1}\mathbf{H}_{:,\mathcal{D}}^{\mathrm{T}} & \mathbf{V}^{-1}\end{bmatrix}\\ &=\begin{bmatrix}\mathbf{Q}^{-1}[\ell] & \mathbf{0}\\ \mathbf{0} & \mathbf{0}\end{bmatrix}+\begin{bmatrix}\mathbf{H}_{:,\mathcal{D}}\\ \mathbf{1}\end{bmatrix}\mathbf{V}^{-1}\begin{bmatrix}\mathbf{H}_{:,\mathcal{D}}^{\mathrm{T}} & \mathbf{1}\end{bmatrix}\end{aligned} \tag{54}$$

where $\mathbf{V}$ and $\mathbf{H}$ are matrices computed based on

$$\begin{cases}\mathbf{H}_{:,\mathcal{D}}=-\mathbf{Q}^{-1}[\ell]\mathbf{Q}_{:,\mathcal{D}}\\ \mathbf{V}=\mathbf{Q}_{\mathcal{D},\mathcal{D}}-\mathbf{Q}_{:,\mathcal{D}}^{\mathrm{T}}\mathbf{Q}^{-1}[\ell]\mathbf{Q}_{:,\mathcal{D}}\end{cases}. \tag{55}$$

For multiple decremental analyses, recall that $\mathcal{R}$ denotes the set of existing training samples that are about to be removed. We can rearrange the elements in $\mathbf{Q}^{-1}$, so that $\mathcal{R}$ lies at the bottom-right corner of $\mathbf{Q}^{-1}$. Let $\mathbf{\Lambda}$, $\mathbf{h}_{\mathcal{R}}$, and $\mathbf{v}_{\mathcal{R}}$ respectively specify the four blocks of $\mathbf{Q}^{-1}$, shown in (56). Besides, $\mathbf{\Lambda}$, $\mathbf{h}_{\mathcal{R}}$, and $\mathbf{v}_{\mathcal{R}}$ represent matrices. Then,

$$\begin{aligned}\mathbf{Q}^{-1}[\ell+1]&=\begin{bmatrix}\mathbf{\Lambda} & \mathbf{h}_{\mathcal{R}}\\ \mathbf{h}_{\mathcal{R}}^{\mathrm{T}} & \mathbf{v}_{\mathcal{R}}\end{bmatrix}\\ &=\begin{bmatrix}\mathbf{Q}^{-1}[\ell]+\mathbf{B}_{:,\mathcal{R}}\mathbf{V}^{-1}\mathbf{B}_{:,\mathcal{R}}^{\mathrm{T}} & \mathbf{B}_{:,\mathcal{R}}\mathbf{V}^{-\mathrm{T}}\\ \mathbf{V}^{-1}\mathbf{B}_{:,\mathcal{R}}^{\mathrm{T}} & \mathbf{V}^{-1}\end{bmatrix}.\end{aligned} \tag{56}$$

Comparing the four blocks in the upper and lower parts of (56) [5] yields the following result.

$$\begin{aligned}&\mathbf{Q}^{-1}[\ell+1]\\ &=\mathbf{\Lambda}-\mathbf{h}_{\mathcal{R}}\mathbf{v}_{\mathcal{R}}^{-1}\mathbf{h}_{\mathcal{R}}^{\mathrm{T}}\end{aligned}. \tag{57}$$

To integrate multiple incremental/decremental processes together, we have

$$\mathbf{Q}^{-1}[\ell+1]=\begin{bmatrix}\mathbf{\Lambda}-\mathbf{h}_{\mathcal{R}}\mathbf{v}_{\mathcal{R}}^{-1}\mathbf{h}_{\mathcal{R}}^{\mathrm{T}}+\mathbf{H}_{:,\mathcal{D}}\mathbf{V}^{-1}\mathbf{H}_{:,\mathcal{D}}^{\mathrm{T}} & \mathbf{H}_{:,\mathcal{D}}\mathbf{V}^{-\mathrm{T}}\\ \mathbf{V}^{-1}\mathbf{H}_{:,\mathcal{D}}^{\mathrm{T}} & \mathbf{V}^{-1}\end{bmatrix}. \tag{58}$$

## Acknowledgement

Dr. Chen would like to thank Prof. Sun-Yuan Kung for his supervision and teaching when Dr. Chen worked as a postdoctoral fellow with Princeton University, USA. Without him, nothing can be done in this study.

## References

[1] S. Li, L. D. Xu, and X. Wang, "Compressed sensing signal and data acquisition in wireless sensor networks and Internet of Things," *IEEE Transactions on Industrial Informatics*, vol. 9, no. 4, pp. 2177–2186, Nov. 2013.

[2] G. Cauwenberghs and T. Poggio, "Incremental and decremental support vector machine learning," in *Proc. 14th Annual Conf. Neural Information Processing System (NIPS)*, Denver, Colorado, United States, 2000, Nov. 28–30, pp. 409–415.

[3] J. C. Platt, "Fast training of support vector machines using sequential minimal optimization," in *Advances in Kernel Methods: Support Vector Learning*, B. Schölkopf, C. J. C. Burges, and A. J. Smola, Eds. Cambridge, Massachusetts, United States: MIT Press, 1999.

[4] C. P. Diehl and G. Cauwenberghs, "SVM incremental learning, adaptation and optimization," in *Proc. International Joint Conference on Neural Networks (IJCNN 2003)*, Portland, Oregon, 2003, Jul. 20–24, pp. 2685–2690.

[5] P. Laskov, C. Gehl, S. Krüger, and K.-R. Müller, "Incremental support vector learning: Analysis, implementation and applications," *Journal of Machine Learning Research*, vol. 7, pp. 1909–1936, 2006.

[6] M. Karasuyama and I. Takeuchi, "Multiple incremental decremental learning of support vector machines," *IEEE Transactions on Neural Networks*, vol. 21, no. 7, pp. 1048–1059, Jul. 2010.

[7] S.-Y. Kung, *Kernel Methods and Machine Learning*. Cambridge, UK: Cambridge University Press, Jun. 2014.

[8] M. Martin, "On-line support vector machine regression," in *Proc. 13th European Conference on Machine Learning*, Helsinki, Finland, 2002, Aug. 19–23, pp. 282–294.

[9] J. Ma, J. Theiler, and S. Perkins, "Accurate on-line support vector regression," *Neural Computation*, vol. 15, no. 11, pp. 2683–703, Nov. 2003.

[10] L. Ralaivola and F. d'Alché-Buc, "Incremental support vector machine learning: A local approach," in *Proc. International Conference on Artificial Neural Networks (ICANN 2001)*, Vienna, Austria, 2001, Aug. 21–25, pp. 322–330.

[11] G. Fung and O. L. Mangasarian, "Incremental support vector machine classification," in *Proc. 2nd SIAM International Conference on Data Mining*, Arlington, Virginia, United States, 2002, Apr. 11–13, pp. 247–260.

[12] A. Tveit and H. Engum, "Parallelization of the incremental proximal support vector machine classifier using a heap-based tree topology," in *Proc. Parallel and Distributed Computing for Machine Learning*, Cavtat-Dubrovnik, Croatia, 2003, Sep. 22–26.

[13] G. Fung and O. L. Mangasarian, "Proximal support vector machine classifiers," in *Proc. 7th ACM International Conference on Knowledge Discovery and Data Mining (SIGKDD 2001)*, San Francisco, California, United States, 2001, Aug. 26–29, pp. 77–86.

[14] H. Drucker, C. J. C. Burges, L. Kaufman, A. J. Smola, and V. Vapnik, "Support vector regression machines," in *Proc. 10th Annual Conference on Neural Information Processing Systems (NIPS)*, Denver, Colorado, United States, 1996, Dec. 3–5, pp. 155–161.

[15] A. J. Smola and B. Schölkopf, "A tutorial on support vector regression," *Statistics and Computing*, vol. 14, no. 3, pp. 199–222, Aug. 2004.

[16] W. H. Press, S. A. Teukolsky, W. T. Vetterling, and B. P. Flannery, *Numerical Recipes 3rd Edition: The Art of Scientific Computing*, 3rd. ed. Cambridge, United Kingdom: Cambridge University Press, 2007.